\documentclass[acmsmall, nonacm]{acmart}
\usepackage{tikz}
\usetikzlibrary{positioning, arrows.meta}
\usepackage{longtable,array}

\setcopyright{none}
\renewcommand\footnotetextcopyrightpermission[1]{} % removes footnote with conference information in first column
\makeatletter
\let\@authorsaddresses\@empty
\makeatother

\AtBeginDocument{%
  }

\begin{document}

\title[Simulating How Adults with Different Skills Evaluate Canada’s National AI Strategy]{Conditional Endorsement: Simulating How Adults with Different Skills Evaluate Canada's National AI Strategy}

\begin{abstract}
  Canada's 2026 national AI strategy, \emph{AI for All}, commits to free AI training for everyone and to expanding business AI adoption, yet it is difficult to anticipate whether these policies will reach adults with uneven skills and circumstances. We ask how adults with diverse skills might rate these policies on support, effectiveness, and personal benefit, how their skills and circumstances shape these evaluations, and what implementation conditions they raise. We built 3,200 generative agents from a weighted sample of the \emph{Programme for the International Assessment of Adult Competencies} (PIAAC) Cycle 2 respondents, collected their ratings and explanations for 20 policies, and analyzed dialogues between agents with contrasting ratings. The agents supported nearly every policy but were far less confident that they would improve adult AI literacy or benefit them personally. This gap widened as digital technology use declined: agents modeled on adults who had never used a digital device still endorsed the policies but rarely expected to benefit. Agents traced the gap to participation costs, hard-to-exercise rights, and limited community control, and their dialogues specified these conditions rather than resolving disagreement. We describe this pattern as \emph{conditional endorsement} and argue that AI literacy policy should be designed and evaluated for adults with the fewest digital skills.
\end{abstract}

\author{Dipto Das}
\authornote{The first author worked on this project while working at the University of Toronto.}
\affiliation{
  \department{College of Computing and Information}
  \institution{Cornell University}
  \city{Ithaca}
  \state{New York}
  \country{United States}
}
\author{F\'elix Leduc}
\affiliation{
  \department{Faculty of Information}
  \institution{University of Toronto}
  \city{Toronto}
  \state{Ontario}
  \country{Canada}
}
\author{Shion Guha}
\affiliation{
  \department{Faculty of Information}
  \institution{University of Toronto}
  \city{Toronto}
  \state{Ontario}
  \country{Canada}
}

\maketitle
%TC:endignore
\section{Introduction}\label{sec:intro}
In June 2026, the Government of Canada released its national artificial intelligence (AI) strategy, \emph{AI for All}~\cite{ised2026aiforall}, that commits to free AI training for everyone in Canada and to raising business AI adoption to 60\% by 2034~\cite{white2026iapp}. Only 12.2\% of Canadian businesses reported using AI in mid-2025~\cite{statcan2025ai}, and almost half of workers who used AI tools at work had received no formal training~\cite{environics2025ai}. For employers, workforce developers, and policymakers, the success of such commitments depends on whether they reach adults with very different skills, jobs, and circumstances. However, before implementation, it is difficult to know how these adults would receive a policy. In this paper, we treat adults' skills and circumstances as drivers of how they receive AI policy, and use generative agents grounded in Canadian adult-skills data to simulate how diverse adults evaluate the strategy's policies.

The problem is hard because the adults whom AI literacy policy most needs to reach are often the least visible in the processes that shape it. Research on the digital divide shows that inequalities in skills and outcomes persist beyond access to devices~\cite{vandijk2020digital, hargittai2002second}. However, national AI strategies tend to articulate public skills and inclusion through broad commitments, while rarely outlining specified mechanisms~\cite{radu2021steering, fatima2020national}. The strategy's public consultation relied on self-selected participation, and critics argued that the strategy was introduced ``without any public debate"~\cite{docto2026dailyhive}.

Large language model (LLM)-based simulations of survey respondents can approximate aggregate opinions~\cite{argyle2023out, park2024generative}. However, concerns remain about misalignment with respondents' views~\cite{santurkar2023whose, brodeur2025comparing} and the flattening of identity groups~\cite{wang2025large}. The Programme for the International Assessment of Adult Competencies (PIAAC)\footnote{Cycle 1: 2011--2017 and Cycle 2: 2022--2023} provides a basis for grounding these simulations in observed variation: it records foundational skills and circumstances for a nationally representative sample of adults~\cite{oecd2024skills}. We use data from its Canadian respondents to construct generative agents as structured, albeit imperfect, proxies for this variation. These agents rate and explain their support for the strategy's policies, their perceived effectiveness, and their anticipated personal benefit. We ask the following research questions:

\begin{itemize}
\item[\textbf{RQ1:}] How do PIAAC-grounded agents evaluate the policies in the national AI strategy?
\item[\textbf{RQ2:}] How do the underlying personas' skills and circumstances shape these evaluations?
\item[\textbf{RQ3:}] What implementation conditions do the agents raise individually and in simulated dialogues?
\end{itemize}

To answer these questions, we constructed 3,200 personas from a weighted PIAAC Cycle 2 sample. Agents rated 20 policies from \emph{AI for All}, yielding over 63,000 valid ratings on each of three dimensions. We then simulated 1,533 dialogues between agents with contrasting ratings. We compared ratings across persona characteristics and analyzed the responses and dialogues with inductive concept analysis~\cite{lam2024concept}. For RQ1, 99.5\% of support ratings fell in the upper half of the scale, compared with 56.4\% of benefit and 35.9\% of effectiveness ratings. For RQ2, expected benefit fell with digital technology use: 63.4\% of benefit ratings from personas with higher use fell in the upper half, compared with 47\% for lower use and 12.3\% for personas who had never used a digital device. For RQ3, the agents attributed the gap to participation costs, rights that are difficult to exercise, and control over resources and institutions; and their dialogues, instead of resolving disagreements, elaborated these conditions. We describe this pattern as \emph{conditional endorsement}: support for AI literacy policy does not imply it will reach adults with varying levels of digital skills equally without active measures. In the following sections, we review related work (Sec.~\ref{sec:literature_review}), describe our research design (Sec.~\ref{sec:research_design}), present our findings (Sec.~\ref{sec:findings}), and discuss their implications and limitations (Sec.~\ref{sec:discussion}).

\section{Literature Review}\label{sec:literature_review}
Our literature review brings together three bodies of work that frame how to evaluate AI literacy policy before it reaches the public: research on AI literacy and adoption, national AI strategies as instruments of governance, and the use of survey-grounded generative agents to simulate policy evaluation.

\subsection{AI Literacy as a Condition of Equitable AI Adoption}
As AI systems become embedded in work, public services, and everyday communication, scholars have increasingly treated AI literacy as a foundational capacity rather than a specialized technical skill. Long and Magerko define AI literacy as a set of competencies that enables individuals to critically evaluate AI technologies, communicate and collaborate effectively with AI, and use AI as a tool online, at home, and in the workplace~\cite{long2020ai}. Similarly, Ng and colleagues' review organizes AI literacy around knowing and understanding AI, using and applying it, evaluating and creating with it, and attending to its ethical implications~\cite{ng2021conceptualizing}. Earlier efforts to define what learners should know about AI~\cite{touretzky2019envisioning} have since been extended to higher and adult education~\cite{laupichler2022artificial}. These conceptualizations shift attention from whether people can operate an AI tool to whether they can judge when, how, and on whose terms to use it. Such judgment also depends on knowing the limits of one's own skill: people who trust automated systems too readily tend to stop checking their output~\cite{parasuraman2010complacency}, and recent AI literacy measures treat self-efficacy as a competency in its own right~\cite{carolus2023mails}.

However, the capacity to acquire such competencies is unevenly distributed. Research on the digital divide has long shown that inequalities extend beyond physical access to devices and connectivity, encompassing differences in skills, usage, and the outcomes people derive from technology~\cite{vandijk2020digital, hargittai2002second, vandeursen2014digital}, and that digital disadvantage tends to compound existing social disadvantage~\cite{robinson2015digital}. AI adoption risks reproducing these layered inequalities: people with lower foundational skills, limited digital experience, or precarious employment may face both fewer opportunities to learn about AI and greater exposure to its disruptions. While LLMs are affecting a large share of work tasks across occupations~\cite{eloundou2024gpts}, the Organization for Economic Co-operation and Development (OECD) has emphasized that training and worker involvement shape whether AI adoption improves job quality~\cite{oecd2023employment}. Adoption also depends on trust. Global studies of public attitudes, most recently spanning 47 countries, have found that people's willingness to rely on AI is closely tied to their confidence in the institutions that develop, deploy, and regulate it~\cite{gillespie2023trust, gillespie2025trust}. Hence, AI literacy and responsible adoption are not only matters of individual learning, but also of the institutional conditions under which learning, access, and trust become possible.

\subsection{National AI Strategies as Instruments of Literacy and Adoption Governance}
Governments have increasingly articulated these institutional conditions through national AI strategies. Since the first national AI strategies appeared in 2017, including the Pan-Canadian Artificial Intelligence Strategy, scholars have examined these strategies as governance documents that set priorities, allocate resources, and signal which publics the state expects to benefit from AI~\cite{radu2021steering, fatima2020national}. Policy documents frame AI through competing narratives of research capacity, economic competitiveness, and societal challenge~\cite{ulnicane2021framing}, and while ethical guidelines converge on principles such as transparency, justice, and responsibility, they diverge on how to implement them~\cite{jobin2019global}.

In Canada, a review of 84 federal and provincial AI governance initiatives found that most focus on industry, research, and public administration, while few address AI education and training or social and workforce development~\cite{attard2024governance}. Similarly, Das and colleagues' analysis of Canada's federal AI Register suggested that, while the register aims to make government AI use visible, it provides insufficient information for the public to contest that~\cite{das2026bureaucratic}. However, Canada's national AI strategy, released in June 2026, places public capacity more centrally than earlier strategies~\cite{ised2026aiforall}. The strategy is organized around six pillars, including protecting Canadians and safeguarding democracy, empowering Canadians, and powering AI adoption. It commits to free AI training for everyone in Canada and to up to 90,000 AI-related jobs and work placements for young Canadians by 2031~\cite{ised2026aiforall}. The strategy also reports a consultation involving more than 11,000 Canadians. Overall, these commitments position AI literacy and adoption as national policy goals that depend on reaching adults across very different occupational, linguistic, and social circumstances.

While a strategy document describes intended actions and measurable targets, it remains unclear how the people it aims to reach might receive them. Public consultations provide valuable input, but they rely on self-selected participation and are typically conducted before policies are specified at the program level. Consequently, we know comparatively little about how adults with different foundational skills, digital experiences, and life circumstances might evaluate specific AI literacy and adoption policies, and what implementation conditions they would consider necessary.

\subsection{Survey-Grounded Generative Agents as Simulated Policy Evaluators}
Answering this question requires both data that describe adults' skills and circumstances and a method that can elicit evaluations of specific policies from those profiles. The OECD's PIAAC provides the former. Its second cycle assessed literacy, numeracy, and adaptive problem solving among adults aged 16--65, alongside detailed background information on education, employment, technology use, learning, and well-being~\cite{oecd2024skills}. While Canadians scored above the average of participating OECD countries across the three domains~\cite{statcan2024piaac}, these national averages conceal substantial variation among adults, which AI literacy policy must address. Since PIAAC links skills and skill use to employment, learning, and social participation~\cite{oecd2024skills}, it allows skills to be examined as outcomes that a given policy seeks to change and as drivers of how individuals might receive that policy.

Recent advances in LLMs have made it possible to simulate how people with particular backgrounds might respond to questions or situations, prompting debate about whether such simulations can complement, or even substitute for, human participants in social science research~\cite{grossmann2023ai, dillion2023can, anthis2025position}. Park and colleagues introduced generative agents, i.e., LLM-based agents that maintain memory, reflection, and planning to produce believable individual and social behavior~\cite{park2023generative}. Subsequent work has grounded such agents in real individuals' data: agents built from qualitative interviews with 1,052 people replicated participants' General Social Survey responses 85\% as accurately as participants replicated their own answers two weeks later~\cite{park2024generative}. Related studies have used conditioned LLMs as ``silicon samples" of survey populations~\cite{argyle2023out}, replicated classic human-subject experiments~\cite{aher2023using}, and simulated economic decision-makers~\cite{horton2023large}. Ghaffarzadegan and colleagues describe this direction as generative agent-based modeling (GABM), in which LLM-powered agents replace hand-coded decision rules, allowing researchers to simulate how heterogeneous agents reason and interact~\cite{ghaffarzadegan2024generative}; similar designs have simulated social media platforms populated by persona-conditioned agents~\cite{tornberg2023simulating}. Moreover, multi-agent exchanges, in which agents respond to and critique one another, have been used to surface and refine reasoning that a single agent might not produce~\cite{du2024improving}, and networks of LLM agents have been used to simulate opinion dynamics~\cite{chuang2024simulating}.

Generative agents have also been applied to social and policy questions, including macroeconomic activity~\cite{li2024econagent}, universal basic income in a society of more than 10,000 agents~\cite{piao2025agentsociety}, vaccine-hesitancy interventions among census-grounded agents~\cite{hou2025can}, team formation at a simulated job fair~\cite{li2025metaagents}, and diplomatic positions on international events~\cite{peng2025diplomacyagent}. These studies mostly model behavior and its aggregate outcomes rather than how people with different profiles evaluate a specific policy, and even census-grounded agents sometimes acted inconsistently with their assigned demographics~\cite{hou2025can}.

However, scholars have also raised various concerns about treating LLM outputs as substitutes for human respondents. LLMs' opinions can be misaligned with those of many demographic groups~\cite{santurkar2023whose}, synthetic survey responses can be less varied and less stable than human responses~\cite{bisbee2024synthetic}, and LLMs prompted to represent identity groups can misportray and flatten them~\cite{wang2025large}. Persona variables may also explain only a limited share of the variation in subjective judgments~\cite{hu2024quantifying}. In interaction, LLMs tend to agree with their interlocutors~\cite{sharma2024towards} and to converge toward consensus~\cite{chuang2024simulating}, which may suppress the disagreement that deliberation is meant to surface. These critiques suggest that GABM should be treated not as a replacement for public consultation, but as an exploratory instrument whose outputs require careful interpretation. Such interpretation can draw on established practice in computationally assisted qualitative analysis, where machine learning helps researchers explore large text corpora while interpretation remains with the researcher~\cite{muller2016machine, nelson2020computational}. Grounding personas in survey data, eliciting ratings with explanations, and analyzing the resulting texts with LLM-assisted tools such as LLooM~\cite{lam2024concept} extend this practice to agent outputs, making their interpretation systematic and scalable.
\section{Research Design}\label{sec:research_design}
\begin{figure*}[!ht]
\centering
\includegraphics[width=\linewidth]{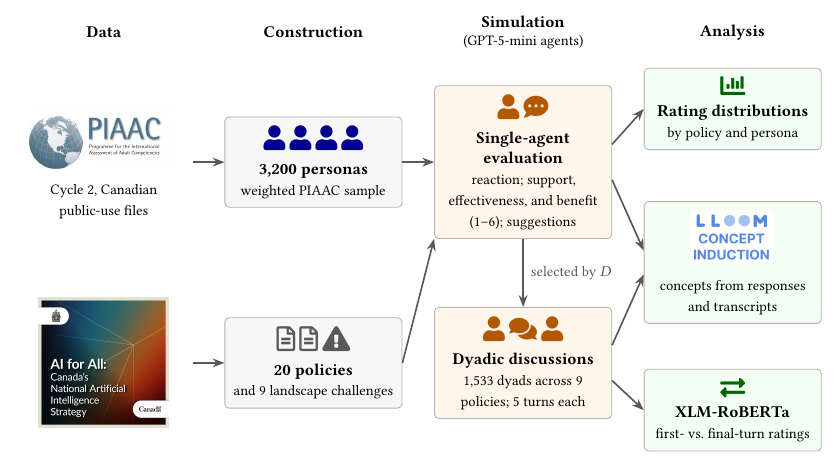}
\caption{An overview of the research design. Personas built from PIAAC responses evaluated the 20 policies identified in \emph{AI for All}, and agents with contrasting ratings on the same policy, selected by the rating distance $D$, then discussed it in dyads. We analyzed the rating distributions, induced concepts from responses and transcripts with LLooM, and used an XLM-RoBERTa model trained on single-agent responses to estimate rating changes within each dialogue.}
\label{fig:design}
\end{figure*}

Our research design combines GABM~\cite{ghaffarzadegan2024generative, park2023generative} with mixed-methods analyses of single-agent responses and simulated dialogues. In this section, we describe the inputs, single-agent evaluation, dyadic discussions, and AI-assisted qualitative analysis. We treat the agents' responses not as estimates of Canadian public opinion, but as simulated, survey-grounded accounts of how people with particular backgrounds, skills, and circumstances might reason about AI policy. Figure~\ref{fig:design} summarizes the research design, and Appendix~\ref{app:methods} provides further procedural details.

\subsection{Policies and Challenges Identification in National AI Strategy}
We conducted a structured qualitative analysis of the \emph{Canada's National Artificial Intelligence Strategy: AI for All} document~\cite{ised2026aiforall}. We coded passages that specified government actions as policies and passages describing risks, barriers, and constraints as landscape challenges. After consolidation, this process yielded 20 policies and nine landscape challenges, such as public trust and AI literacy, unequal access, and privacy and online harms (Appendix~\ref{app:policies}). We used an LLM to paraphrase each policy and challenge into a self-contained description.

\subsection{PIAAC-grounded Persona Construction}
Before constructing agents from PIAAC respondents, we examined the PIAAC data directly. Appendix~\ref{app:overconfidence} compares Canadian adults' ratings of their own assessment performance with their measured proficiency and finds that confidence tracks proficiency only loosely: at equal proficiency, men and adults with a bachelor's degree or higher were more likely to rate their performance above what their scores supported. Because how adults perceive their skills can differ from what PIAAC measures, the personas below express skills relative to the weighted sample rather than through respondents' own self-assessments.

We drew a weighted sample of 3,200 records from the PIAAC Cycle 2 public-use files (PUF)~\cite{oecd2024skills} using the final full-sample weight (Appendix~\ref{app:piaac})\footnote{Based on the sample weights, some PIAAC profiles were sampled for multiple personas.}. Following prior work that conditions LLMs on survey respondents' characteristics~\cite{argyle2023out, park2024generative}, we constructed a synthetic persona from each sampled respondent's selected PIAAC characteristics, including demographic background, language, education, employment, household and economic context, digital technology use, learning and training, personality, social trust, and health. Thus, the personas embodied skills in terms of education, skill use at home and at work, and digital technology use, expressed as lower, typical, or higher levels relative to the weighted sample (Appendix~\ref{app:personas}). These personas are synthetic representations derived from PIAAC survey responses. Each persona was represented by one generative agent, instantiated as a separate instance of GPT-5-mini. We also assigned each persona a randomly selected subset of the landscape challenges, which served as context for considering how broader implementation barriers might shape the persona's assessment.

\subsection{Single-Agent Policy Evaluation}
We presented each agent with the policy announcements and asked it to respond from its persona's perspective, considering the persona's background, skills, circumstances, and likely priorities. For each policy, the agent provided (1) a brief first reaction; (2) three six-point ratings, each with a brief justification, of support for the policy, the policy's perceived effectiveness, and the likelihood that it would directly benefit the persona; and (3) one to three sentences suggesting changes that might make the policy more effective for someone in the persona's circumstances, given the assigned challenges. The effectiveness rating asked how effective the policy would be in improving AI literacy among Canadian adults, and the benefit rating asked how likely the policy was to benefit the persona directly. We generated prompts in English or French based on the persona's PIAAC questionnaire language, and instructed agents not to add personal details absent from their descriptions (Appendix~\ref{app:prompts}).

\subsection{Dyadic Discussions}
For each policy, we distinguished lower-rating agents, who rated a policy 1--3 on support, effectiveness, and benefit, from higher-rating agents, who rated it 4--6 on all three. A contrasting dyad paired one agent from each group for that particular policy. Only 9 of the 20 policies had agents in both groups, yielding 15,222 possible dyads. For each dyad, we calculated the L1 distance across the three ratings:

\begin{equation}
D = |S_{\mathrm{low}} - S_{\mathrm{high}}| + |E_{\mathrm{low}} - E_{\mathrm{high}}| + |B_{\mathrm{low}} - B_{\mathrm{high}}|,
\end{equation}

where $S$, $E$, and $B$ denote support, effectiveness, and benefit ratings. After excluding mixed-language pairs, we selected approximately 10\% of the dyads ($n$ = 1,533) using a coverage-first procedure: up to 10 high-distance dyads per eligible policy, followed by selecting the remaining dyads based on the highest $D$ across policies (Appendix~\ref{app:dyads}).

Then, we asked the two agents in each dyad to discuss the policy on which they had previously provided contrasting ratings. The conversation proceeded in alternating turns, with each agent allowed up to five turns, responding to the other agent's most recent comment, and instructed to contest, qualify, connect, or build on the other speaker's arguments. To characterize how conversations opened, we checked whether the second speaker's first turn began with an explicit agreement phrase (e.g., ``I agree" or ``You're right") in English or French.

To examine whether the initial differences changed during the conversation, we fine-tuned a multilingual XLM-RoBERTa model~\cite{conneau2020unsupervised} on the single-agent responses and their ratings, with one classification head per rating dimension, and evaluated it on held-out response--rating pairs (Appendix~\ref{app:model}). We used the model to predict support, effectiveness, and benefit ratings for each conversational turn and compared each agent's predicted ratings in the first and final turns. We compared these estimates with a TF-IDF logistic-regression classifier trained on the same data (Appendix~\ref{app:corpus}). Based on changes between each agent's first and final turns, we classified dyads as converging, maintaining their initial disagreement, or becoming more polarized.

\subsection{Qualitative Analysis}
We conducted an automated qualitative analysis of the single-agent responses and the dyad conversations using LLooM~\cite{lam2024concept}, which generates concepts inductively from text. For the single-agent responses, we ran LLooM separately for each policy on the first reactions, rating justifications, and suggested changes. For the dyads, we analyzed each transcript as a single document, identifying disagreements, shared positions, and recommendations introduced or developed in later turns (Appendix~\ref{app:lloom}). We used the concepts as probes for interpretation, and not as prevalence estimates. We attended to the explanations and transcript snippets the concepts pointed to, noting how agents invoked their personas' skills and circumstances, and organized the findings into themes that triangulate the ratings, single-agent explanations, and conversations. In the findings, we cite single-agent responses by persona ID and conversational turns in dyads by speaker ID, turn, and partner. Appendix~\ref{app:repro} describes our reproducibility measures.

\section{Findings}\label{sec:findings}
We organize the findings around five themes. The first shows that the agents' near-universal support for the 20 policies did not translate into expectations of literacy reach or personal benefit. The second shows how the personas' skills and circumstances, particularly their digital technology use and employment, shaped who expected to benefit. The third and fourth trace the conditions that the agents attached to their support: participation costs, and rights and control that people could actually exercise. The fifth examines what the dyadic conversations added to these conditions and what they cannot show. Within each theme, we triangulate across the ratings, explanations, and conversations, and we note evidence that complicates the pattern.

\subsection{Near-Universal Support Alongside Divided Effectiveness and Benefit}\label{sec:findings-endorsement}
\begin{figure}[!ht]
\centering
\includegraphics[width=\linewidth]{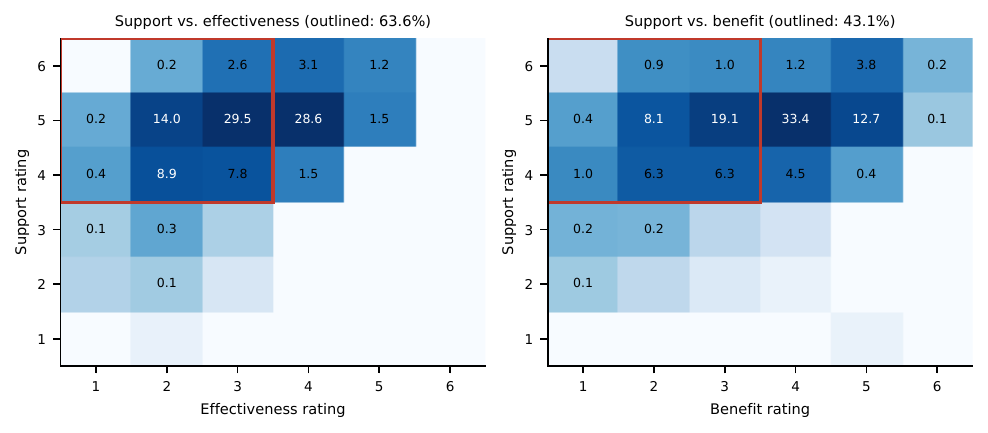}
\caption{Joint distribution of support with literacy effectiveness (left) and direct benefit (right) among the 63,875 responses with all three ratings. Cell values are percentages of these responses (values below 0.05\% omitted), and outlined cells mark upper-half support combined with a lower-half rating.}
\label{fig:crosstab}
\end{figure}

Support, direct-benefit, and effectiveness ratings had markedly different distributions (Figures~\ref{fig:support}--\ref{fig:benefit} in Appendix~\ref{app:figures}). Of the valid ratings, 99.5\% of support ratings fell in the upper half of the scale (4--6), compared with 56.4\% of direct-benefit and 35.9\% of effectiveness ratings. Every policy had more than 95\% of its valid support ratings at 4--6. Within the responses with all three ratings present, 63.6\% combined upper-half support with a lower-half effectiveness rating, and 43.1\% combined upper-half support with a lower-half direct-benefit rating (Figure~\ref{fig:crosstab}).

The rating prompts explain much of this divergence. ``Effectiveness" asked whether a policy would improve AI literacy among Canadian adults, not whether it would achieve all of its stated objectives, and ``benefit" asked about direct benefit to the persona. For example, the \emph{consumer privacy reform} policy received upper-half support from every valid response and upper-half benefit ratings from 98.5\%, yet only 10.7\% rated it in the upper half for improving adult AI literacy (Table~\ref{tab:upperhalf} in Appendix~\ref{app:tables}). Upper-half support exceeded 99\% for an \emph{AI mission in health care}, \emph{health-data research platforms}, and \emph{AI adoption in federal public services}, but at most 0.6\% rated their literacy effectiveness at 4--6. Conversely, effectiveness was highest where a policy had an explicit educational component: all valid ratings for free introductory AI education were 4 or 5, whereas 91.6\% of ratings for financing and commercialization were 1 or 2. Thus, these proposals could be valued as privacy, health, or economic initiatives while being seen as weak literacy interventions.

In turn, direct benefit separated public endorsement from personal proximity. For the \emph{Indigenous leadership in AI} policy, every valid support rating was 4--6, whereas only 0.4\% of direct-benefit ratings were rated in the upper half. For the \emph{AI tools for post-secondary students} policy, the figures were 99.9\% and 6.7\%. The agents located this gap in their own eligibility rather than in the policy's value:

\begin{quote}
``I welcome a policy that centers Indigenous leadership---supporting languages and cultural expression feels important and right to me even though I'm not Indigenous. [...] I'm not Indigenous and I'm out of the labor force, so this would probably not change my day-to-day life even if I appreciate its goals." \hfill (PIAAC-5597)
\end{quote}

\noindent A response about post-secondary tools similarly stated, ``I'm not a student or an educator now, so I'm unlikely to get direct personal benefit from this program" (PIAAC-11312-3). These low benefit scores reflect the item's individual framing and the policy's target population, not a judgment that Indigenous leadership or student access is unimportant.

\subsection{Skills and Circumstances Shaped Who Expected to Benefit}
Because each agent reasoned from a PIAAC-derived profile, we could examine how skills and circumstances operated as drivers of its evaluations. Support barely varied across persona characteristics, but expected benefit and literacy effectiveness did (Table~\ref{tab:drivers}).

\begin{table*}[!ht]
\caption{Share of valid ratings in the upper half of the scale (4--6) across all 20 policies, by persona characteristic. Levels of digital technology use at home are relative to the weighted sample.}
\label{tab:drivers}
\small
\begin{tabular}{llrrr}
\toprule
Characteristic & Level & Support (\%) & Effect. (\%) & Benefit (\%) \\
\midrule
Digital use at home & Never used a device & 90.4 & 13.2 & 12.3 \\
 & Lower & 99.4 & 31.0 & 47.0 \\
 & Typical & 99.7 & 37.5 & 57.7 \\
 & Higher & 99.8 & 39.1 & 63.4 \\
\midrule
Employment status & Out of the labour force & 98.7 & 32.4 & 45.3 \\
 & Unemployed & 99.3 & 37.0 & 52.6 \\
 & Employed & 99.7 & 36.9 & 59.2 \\
\midrule
Education & Below upper secondary & 98.5 & 29.5 & 41.8 \\
 & Upper secondary\textsuperscript{a} & 99.4 & 35.1 & 52.2 \\
 & Tertiary & 99.9 & 38.6 & 63.5 \\
\bottomrule
\end{tabular}
\par\smallskip\footnotesize\textsuperscript{a} Includes post-secondary non-tertiary education (ISCED 3--4).
\end{table*}

\subsubsection{Digital Technology Use as a Lens on Benefit}

Digital technology use showed the steepest gradient. Across the 20 policies, 63.4\% of benefit ratings from personas with higher digital technology use at home fell in the upper half, compared with 47.0\% from those with lower use. The 48 personas who had never used a digital device rated benefit in the upper half only 12.3\% of the time and effectiveness only 13.2\%, although 90.4\% of their support ratings were 4--6. Personas with lower digital use expected less benefit than those with higher use for 19 of the 20 policies, with the largest gaps for the ones about \emph{AI support for Canadian creators} (15.9\% vs.\ 70.1\%) and \emph{support for open-source AI} (36.8\% vs.\ 83.5\%). The agents' explanations show this gradient from both sides. Among benefit rationales for accessibility and inclusion standards, LLooM surfaced a Daily Digital Use concept, in which agents expected some benefit because they already relied on AI-enabled products:

\begin{quote}
    ``I use digital tools every day, so accessible AI would likely make some products I rely on easier and safer" (PIAAC-7748).
\end{quote}

Agents with low digital use reasoned in the opposite direction about public-service automation:

\begin{quote}
``
I'm wary of this---I don't trust new tech in government and worry it will make services harder for people like me who aren't online. [...] It's somewhat unlikely to directly help me since I don't use digital devices and depend on non-digital ways to deal with government offices." \hfill (PIAAC-1754)
\end{quote}

\noindent Similarly, an agent based on the persona of a self-employed craft worker (PIAAC-2950-2) ``with low digital use for transactions and no workplace training" expected no benefit from accessibility standards ``unless there's local, practical help". This pattern complicates the strategy's premise. The adults whom AI literacy policy most needs to reach were the least likely to expect it to reach them, because its delivery appeared to assume the digital engagement they lacked. The one exception was \emph{free introductory AI education}, the policy designed for beginners: personas with lower digital use rated its benefit in the upper half more often than those with higher use (94.8\% vs.\ 82.9\%).

\subsubsection{Employment, Migration, and Health as Frames for Evaluation}
Employment and education produced smaller but consistent gradients. Personas out of the labor force rated benefit in the upper half 45.3\% of the time, compared with 59.2\% for employed personas, and personas without upper secondary education did so 41.8\% of the time, compared with 63.5\% for those with tertiary education. Employment, migration, and health also shaped which features agents valued. Unemployed PIAAC respondents' persona-based agents welcomed faster public-service processing because they ``depend[ed] on services" and could receive ``benefits or approvals sooner," while worrying about losing human help when decisions affected them (PIAAC-6882-3, PIAAC-10565). An agent based on the persona of a first-generation immigrant ``with limited digital skills" doubted that accessibility standards would help without ``plain-language resources, translations, and local outreach" (PIAAC-5825), and an agent whose ``poor health and limited digital comfort" could leave them ``left out of new tech quickly" made a similar demand (PIAAC-8440). Thus, the same policy was evaluated through different stakes: speed of access for those who rely on services, usability for those already online, and the risk of exclusion for those with limited digital skills.

\subsection{Participation Required Time, Devices, and Support}
Agents treated formal eligibility for funding, training, or tools as only the first condition of participation.

\subsubsection{Access Beyond Formal Availability}
In the \emph{support for small-business AI adoption} policy, one agent wrote that ``money alone won't fix" lack of expertise, fragmented data, vendor lock-in, and weak worker training (PIAAC-1258). Another worried that the funds might reach better-connected firms while frontline workers experienced job changes without training (PIAAC-1409). The LLooM concepts of Paid Training Time, Partnerships with Colleges, and Childcare and Support Services reflect the same constraints: a place in a course, a loan, or an AI account was not sufficient if attendance cost wages, childcare, travel, or device access, echoing prior scholarship that locates digital inequality in skills and usage~\cite{vandeursen2014digital, robinson2015digital}. In a post-secondary-tools dyad, a lower-rating agent challenged the assumption that an AI tool could serve as its own entry point:

\begin{quote}
PIAAC-10138-3: ``Use the AI agents themselves to help bridge the gap: have them run plain-language, step-by-step on-boarding [...] tied to local supports like libraries, device-lending and community mentors so people get guided practice, not just a login."
\par\smallskip
PIAAC-1799-2: ``[...] for people like me who've never used a device those digital on-boarding bots come too late. Before an AI can help, there must be local, in-person supports---device lending, short hands-on sessions at libraries or community centers, and a phone helpline---plus printed step-by-step guides I can keep."
\par\smallskip
PIAAC-10138-3: ``[...] I'd add two practical ideas: recruit and train student `digital ambassadors' to run short pop-up sessions at libraries, community centers and campuses [...], and pre-configure loaned devices with an offline step-by-step tutorial [...] so newcomers can try things without needing data or a complex sign-up."
\end{quote}

\noindent Thus, the higher-rating agent revised its own proposal, moving the AI agent from the entry point to a step that follows in-person support.

\subsubsection{Participation Costs as Program Design Requirements}
Recommendations made these costs part of program design. One employment response proposed ``short accredited, hands-on modules delivered by local colleges or unions," paid time off, and ``portable credentials" (PIAAC-11035-5), and a self-employed agent asked for ``flexible, low-cost micro-credentials and remote short courses" (PIAAC-10001). A dyad developed a related chain of supports, from paid training blocks and childcare or travel help to community ``on-ramp" programs with device loans (PIAAC-3906-3 and PIAAC-10008-4, turn 0). By the final turn, the discussion had reached an issue absent from a simple enrollment target: training stipends might interfere with social assistance. The partner proposed a written assessment of benefit eligibility, so participants would not be ``forced to choose between retraining and basic needs" (PIAAC-10008-4, turn 4). This progression locates exclusion in the interaction among programs, not only in individual willingness to learn. Small-business related dyads similarly translated adoption funding into service obligations, such as on-site coaches, vendor setup visits, a human helpline, and low-risk pilot tasks (PIAAC-11278, turn 0, paired with PIAAC-10216-5), although they generated many plausible features without weighing cost or choosing among them.

\subsection{Rights and Control Had to Be Exercisable}
Though the policies that were committed to Privacy, oversight, and community-leadership drew support, agents questioned whether people could actually use them, who would govern resources, and which institutions the proposed safeguards would constrain.

\subsubsection{Consent and Oversight as Supported, Contestable Processes}
Across policies, the issue was less a generalized distrust of AI than uncertainty about who could understand, contest, or enforce the promised protections. One agent worried about ``who can access the data and whether privacy promises will really hold" (PIAAC-4809-2), and another feared that ``real people could lose control over important decisions without clear ways to appeal" (PIAAC-9305). In a dyad focused on health data, one agent made the non-digital route concrete:

\begin{quote}
``A non-digital opt-out/control must be real: paper forms at clinics, a staffed phone line, and community-center assistance in multiple languages. Funded in-person navigators (at hospitals and public health offices) to explain uses and help people opt out or consent would make this usable for older, non-tech-savvy patients." (PIAAC-5595, turn 0, paired with PIAAC-1799-2)
\end{quote}

\noindent Subsequent turns added independent patient representation, record correction, revocable consent, and mailed or telephone confirmation. Thus, consent was treated not as a one-time notice but as a process that had to remain understandable, reversible, and supported by someone accountable, echoing critiques of notice-and-consent regimes~\cite{solove2013privacy}. The public-service dyad followed a comparable path. The lower-rating speaker asked who decides which automated decisions are ``meaningful" (PIAAC-1799-2, turn 0, paired with PIAAC-1532), and the partner proposed a threshold based on consequences for legal status, benefits, or essential services, along with documented human review, paper or in-person options, and pilots with published error and exclusion rates. The promise of human oversight thereby became a sequence of decisions about who qualifies for review, how a person learns what happened, and what remedy follows an error.

One low-rating persona (PIAAC-1799-2) recurred across several of these dyads, which complicates any claim that the dialogues independently converged on non-digital routes. However, single-agent suggestions raised the same routes without a partner, calling for ``local help (phone or in-person) for people not comfortable with digital systems" (PIAAC-2952-3) and for ``non-digital service options so people with low digital skills or language barriers aren't left behind" (PIAAC-2711-2).

\subsubsection{Community Control over Funds and Knowledge}
In the creator-policy responses, one agent worried that the program could miss Indigenous and racialized creators or neglect consent (PIAAC-3067-2), and another proposed reporting ``who receives funding and how datasets are used" (PIAAC-3127). Indigenous leadership responses asked for funding that is ``multi-year and controlled by communities" (PIAAC-1562-2) and safeguards for ``Indigenous data sovereignty and IP" (PIAAC-3349). These claims concern not only entry into a program but authority to govern knowledge, data, and the terms of participation, consistent with Indigenous data governance principles~\cite{carroll2020care, fnigc2014ocap}. The creator dyad extended this authority to withdrawal, moving from in-person help and template contracts to consent that could expire, with the partner's final turn proposing:

\begin{quote}
PIAAC-1799-3: ``Add this for people like me who can't read long letters or go online: require a mailed, plain-language, signed certificate of deletion plus a phone call from a local liaison to confirm action. Let people name a trusted proxy (family member, community rep) to act and receive confirmations on their behalf."
\par\smallskip
PIAAC-5205: ``[...] two practical things: require an alternative non-written confirmation (a recorded phone call or short audio certificate sent on request) for people who can't read, and give creators a simple time-limited consent option (or automatic expiry/auto-delete after a set period) so they don't feel locked in forever."
\end{quote}

\subsubsection{Safeguards as Constraints on Enforcers and Dominant Firms}

Other concerns targeted the institutions responsible for a policy. An online-safety respondent asked for ``transparency and appeal mechanisms so enforcement doesn't unintentionally silence legitimate speech" (PIAAC-8328). In a dyad on international technology cooperation, a lower-rating agent argued that dominant firms already had ``the money and contracts to shape standards and win procurement" (PIAAC-1799-2, turn 0, paired with PIAAC-6187-3), and the partner translated this concern into procurement criteria for interoperability, independent testing, and public progress reporting. These proposals would change the obligations of governments, vendors, and platforms rather than individual AI literacy. The dyad later wandered into detailed procedures for home visits, illustrating how the instruction to keep adding something new could pull a conversation away from the policy's core mechanism.

\subsection{Dialogue Specified Delivery Conditions but Did Not Resolve Disagreement}
% \subsubsection{Contrasting Ratings, Cooperative Openings}
Although every dyad paired an agent who rated the policy 1--3 on all three dimensions with one who rated it 4--6, the contrast rarely produced an adversarial opening. The second speaker began with an explicit agreement phrase in 91.6\% conversations. In one small-business policy-related dyad, the first agent's prior ratings were 3, 2, and 1, whereas the partner's were 5, 4, and 5. Their first turns nonetheless began as follows:

\begin{quote}
    PIAAC-11278: ``I welcome the money---but as someone who's never used a digital device, I worry loans and tools alone won't get small businesses to adopt AI. For many older owners or those in poor health, the barrier is knowing where to start and having patient, hands-on help."
    \par\smallskip
    PIAAC-10216-5: ``I agree---money and generic tools won't be enough for many small firms. From my experience, practical, on-site help makes the difference: sending trained `AI coaches' to a shop or office, running short hands-on workshops at local libraries or community colleges."
\end{quote}

\begin{figure*}[!ht]
\centering
\begin{tikzpicture}[
  font=\scriptsize,
  head/.style={align=flush center, text width=38mm, font=\scriptsize\bfseries},
  concern/.style={draw, rounded corners, fill=gray!20, align=flush center, text width=36mm, minimum height=9mm, inner sep=1.5mm},
  mech/.style={draw, rounded corners, align=flush center, text width=36mm, minimum height=7mm, inner sep=1.5mm},
  arr/.style={-{Stealth[length=1.6mm]}, thick}]
\foreach \x/\title/\ids/\c/\ma/\mb/\mc/\md in {
  0mm/{Small-business adoption}/{PIAAC-11278 and PIAAC-10216-5}/{``Loans and tools alone won't get small businesses to adopt AI"}/{On-site AI coaches and hands-on workshops}/{Basic device training and a phone support line}/{Vetted tools, vendor warranties, and a right to revert}/{Vendor setup on site with follow-up visits},
  45mm/{Health data}/{PIAAC-1799-2 and PIAAC-5595}/{``I worry about the privacy promise"}/{Paper forms, a staffed phone line, and community-centre help}/{Independent oversight with patient representatives}/{In-person record correction and appeals}/{Revocable consent confirmed by mail or phone},
  90mm/{Federal public services}/{PIAAC-1799-2 and PIAAC-1532}/{``Who decides what's `meaningful'?"}/{Impact-based threshold for human review}/{Mandatory pilot publishing error and exclusion rates}/{Oversight board with authority to halt systems}/{Mailed plain-language notice before automation}} {
  \node[head] (h) at (\x,0) {\title\\[-1pt]{\normalfont\tiny \ids}};
  \node[concern, below=2mm of h] (c) {Turn~0 concern:\\ \c};
  \node[mech, below=4mm of c] (m1) {\ma};
  \node[mech, below=3mm of m1] (m2) {\mb};
  \node[mech, below=3mm of m2] (m3) {\mc};
  \node[mech, below=3mm of m3] (m4) {\md};
  \draw[arr] (c) -- (m1);
  \draw[arr] (m1) -- (m2);
  \draw[arr] (m2) -- (m3);
  \draw[arr] (m3) -- (m4);
}
\end{tikzpicture}
\caption{Acknowledgment followed by specification in three dyads. Each column runs from the lower-rating speaker's opening concern to mechanisms the two speakers added in later turns.}
\label{fig:chains}
\end{figure*}
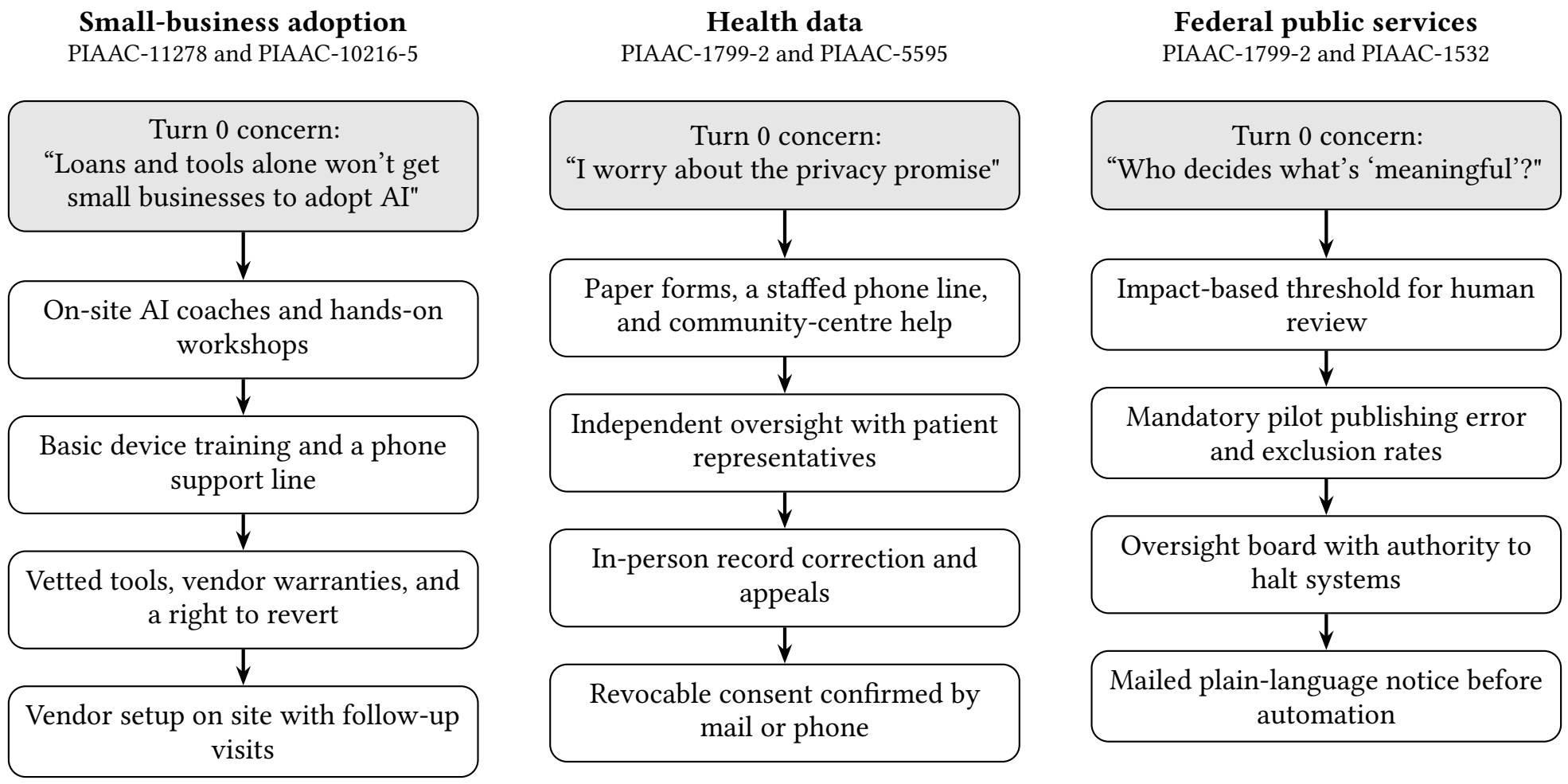

% \subsubsection{Acknowledgment Followed by Specification}
Across topics, the recurring conversational move was acknowledgment followed by specification (Figure~\ref{fig:chains}). In an exchange about the small-business focused policy, ``money and generic tools won't be enough" led to requests for on-site coaches, vendor follow-up, and a person who could troubleshoot. In dyads regarding health data, a broad privacy concern generated recommendations for paper forms, staffed lines, independent patient representatives, and revocation confirmation. In dyads focused on public services, a question about ``meaningful" oversight generated criteria for human review, measurable pilots, appeals, and remediation. These chains converge with the single-agent suggestions reported above, which raised the same mechanisms without a partner.

% \subsubsection{What the Conversations Cannot Show}
We did not ask agents to report a new rating in the middle of their conversations, so opinion changes were inferred from the first and last turns, and the estimates diverged by method. One model classified support as unchanged for 98.4\% of the speaker trajectories and another for 86.2\% (Appendix~\ref{app:corpus}). As the agents were instructed to ``address the other speaker's most recent point" and to ``add a new perspective instead of repeating [their] entire position,", the setup likely encouraged elaboration over conflict. Overall, the conversations document how agents turned a policy's abstract promise into candidate delivery conditions with little evidence of persuasion.

Our analysis highlights why endorsement for the policies in the national AI strategy alone is an incomplete measure of policy acceptance: a public aim could be supported while interrogating and doubting who would reach the service, who could govern its use, and what recourse would exist when its safeguards failed.
\section{Implications and Future Directions}\label{sec:discussion}
Our findings suggest that PIAAC profiles-grounded agents evaluated Canadian AI policies along at least three distinct axes: whether a policy's public aim deserved support, whether it would reach adults as an AI literacy intervention, and whether it would reach the persona personally. Near-universal support coexisted with much lower effectiveness and benefit ratings, a gap that widened as the personas' digital technology use, employment, and education declined, and that the agents located in the conditions of delivery rather than in the policies' value. Below, we discuss our contributions, implications, and limitations.

\subsection{Conditional Endorsement as a Lens on AI Policy Evaluation}
\begin{figure*}[!ht]
\centering
\begin{tikzpicture}[
  box/.style={draw, rounded corners, align=flush center, text width=24mm, minimum height=12mm, font=\small},
  cond/.style={draw, rounded corners, align=flush left, text width=46mm, minimum height=10mm, font=\small, fill=gray!10},
  note/.style={align=flush center, text width=38mm, font=\footnotesize\itshape},
  arr/.style={-{Stealth[length=2mm]}, thick}]
\node[cond] (rights) {\textbf{Exercisability}: non-digital channels, revocable consent, human review and appeal};
\node[cond, above=4mm of rights] (access) {\textbf{Participation}: paid training time, devices, in-person help, benefit eligibility};
\node[cond, below=4mm of rights] (control) {\textbf{Control}: community-governed funds and data, constraints on enforcers and vendors};
\node[box, left=9mm of rights] (promise) {\textbf{Policy promise}\\ \footnotesize public aim};
\node[box, right=9mm of rights] (reach) {\textbf{Reach and benefit}\\ \footnotesize literacy gains, personal relevance};
\node[note, above=18mm of promise] (s) {Measured by support};
\node[note, above=18mm of reach] (e) {Measured by effectiveness and benefit};
\draw[dotted] (s) -- (promise);
\draw[dotted] (e) -- (reach);
\foreach \c in {access, rights, control} {
  \draw[arr] (promise.east) -- (\c.west);
  \draw[arr] (\c.east) -- (reach.west);
}
\node[box, text width=48mm, below=8mm of control] (dyad) {\textbf{Dyadic dialogue}: acknowledgment $\rightarrow$ specification of delivery conditions};
\draw[arr, dashed] (dyad.north) -- (control.south);
\node[box, fill=gray!10] (skills) at (reach |- dyad) {\textbf{Skills and circumstances}\\ \footnotesize digital use, employment, education};
\draw[arr, dashed] (skills.north) -- (reach.south);
\end{tikzpicture}
\caption{Conditional endorsement. Agents endorsed a policy's public aim but qualified whether it would reach adults, or the persona, through three delivery conditions. The personas' skills and circumstances moderated expected reach and benefit, and dyadic dialogue mainly specified the delivery conditions.}
\label{fig:conditional}
\end{figure*}
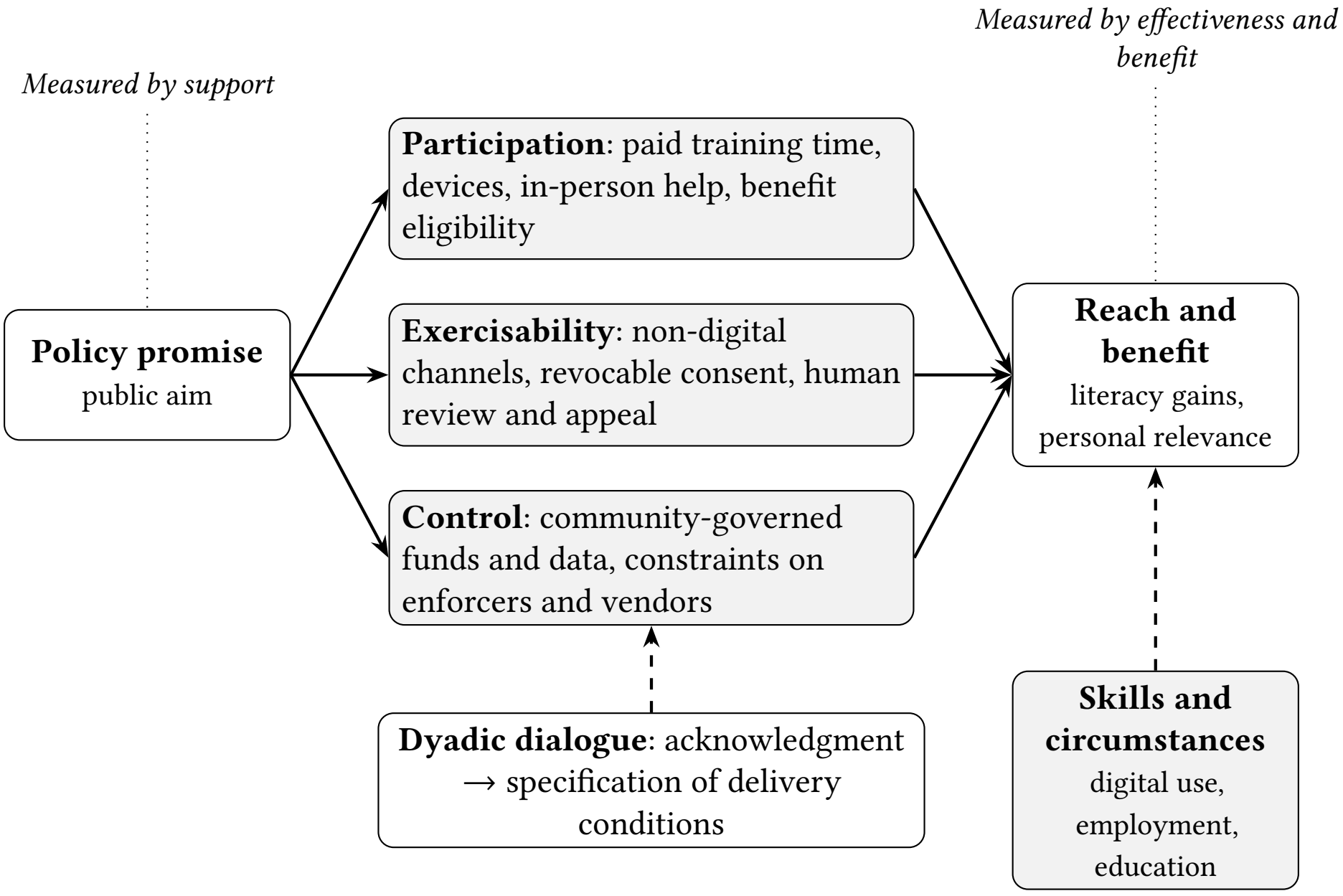

Based on the synthetic policy evaluations by PIAAC-grounded agents, we propose an AI policy evaluation lens, which we dub \emph{conditional endorsement}: support for a policy's public aim that is contingent on how the policy is delivered, who can exercise it, and who governs it. Therefore, a high support rating is not a reliable verdict on whether a policy will work for a given person, but an endorsement qualified by three delivery conditions (Figure~\ref{fig:conditional}): participation, i.e., the time, devices, and people that attendance requires; exercisability, i.e., whether rights and oversight can be used through ordinary channels; and control, i.e., whether affected communities govern resources and safeguards constrain enforcers and vendors. These conditions bind unevenly: because they presuppose time, devices, and digital fluency, they weigh most on adults with the fewest digital skills, for whom expected benefit become minimal while support can still stay high. Hence, conditional endorsement becomes a framework for interpreting where, and for whom, an endorsed policy's promise may fail to translate into reach and benefit.

Our study also makes a methodological contribution. By pairing survey-grounded personas with separate ratings for support, effectiveness, and benefit, the design makes visible a gap that a single approval measure would obscure, and shows how that gap is distributed across adults' skills. Moreover, the dyadic conversations functioned less as a persuasion experiment~\cite{chuang2024simulating, anthis2025position} than as a device for specification, turning abstract concerns into delivery requirements.

\subsection{Implications for AI Literacy Policy and Practice}
First, policymakers and evaluators should not treat high support as evidence of effective or equitable AI literacy policy. A privacy reform can be valued while being seen as a weak literacy intervention, and a policy for Indigenous communities can be endorsed by individuals who do not expect to benefit. Consultation instruments should measure endorsement, reach, and personal relevance separately, report them together, and break them down by skill level. As public reactions questioned whether the national AI strategy would benefit ordinary Canadians~\cite{docto2026dailyhive}, those distinctions also need to be contextualized in the public socioeconomic conditions and sociotechnical practices.

Second, AI literacy programs should be designed first for adults with little or no digital experience. Free introductory AI education was the only policy that personas with lower digital use expected to benefit from more often than those with higher use, which suggests that explicitly beginner-oriented programs can reverse the gradient that other policies reproduced. Such programs should offer in-person, telephone, and print entry points rather than treating an AI tool as its own onboarding. Programs should also treat participation costs as design requirements rather than as individual barriers. Almost half of Canadian workers who used AI tools at work had received no formal training~\cite{environics2025ai}, so training cannot be assumed to follow from access to tools. Employers can contribute paid training time and on-site coaching, and service providers can offer device lending, childcare or travel support, and in-person help. Designers should also check that training stipends do not jeopardize social assistance.

Designing for adults with the fewest digital skills addresses reach, but using AI well also requires knowing when to check its output. A supplementary analysis of Canadian PIAAC Cycle~2 respondents (Appendix~\ref{app:overconfidence}) suggests that the adults most likely to overestimate their own skills are not those with the fewest digital skills. Among adults who completed PIAAC's self-assessment item, men with a bachelor's degree were the most likely to rate their performance above what their measured proficiency supported (about 14.6\%), whereas women aged 55 and over with lower secondary schooling or less were the least likely (5.3\%; Table~\ref{tab:oc-personas}). Because over-confidence of this kind can discourage people from verifying AI outputs~\cite{parasuraman2010complacency}, beginner-oriented programs could be complemented by verification and calibration content aimed at confident, credentialed adults, who are unlikely to see themselves as the audience for introductory AI training. This analysis uses self-assessment on the PIAAC assessment as a proxy and does not measure over-trust in AI directly.

Third, privacy and oversight commitments must be exercisable by people with limited digital fluency. This requires paper and telephone channels, plain-language and multilingual notices, revocable consent, explicit thresholds for human review, and published error and exclusion rates, so that oversight can be contested rather than merely promised. Prior work cautions that human oversight requirements can legitimize flawed systems without offering meaningful protection~\cite{green2022flaws}, and that transparency instruments can provide visibility without contestability~\cite{das2026bureaucratic, lyons2021conceptualising}.

Fourth, community leadership and institutional safeguards require control rather than consultation alone, as Indigenous data governance principles and international guidance on AI ethics emphasize~\cite{carroll2020care, fnigc2014ocap, unesco2021recommendation}. Community-led AI initiatives should provide multi-year funding controlled by communities and protect data sovereignty, whereas procurement and online-safety rules should constrain dominant firms and enforcers through open standards, independent testing, and appeal mechanisms.

\subsection{Limitations and Future Work}
This study has several limitations. First, the personas are synthetic. While simulations of this kind can help policymakers anticipate where delivery may break down, they cannot replace consultation with Indigenous communities, displaced workers, or people without digital access, whose own accounts remain necessary to validate and prioritize the conditions the agents raised. The generative agents' responses should be interpreted as Canadian public's survey-grounded simulated reasoning. This caveat applies with particular force to the skill gradients: because the persona descriptions stated characteristics such as digital technology use explicitly, the gradients may partly reflect the model's assumptions about these labels. The gradients are also descriptive, rely on skill use rather than assessed proficiency, and rest in places on small groups, such as the personas who had never used a digital device. Appendix~\ref{app:overconfidence} draws on PIAAC's assessed proficiency directly, but only to examine how adults rate their own skills, not how they evaluate the policies. Second, the rating items were narrowly framed: effectiveness concerned adult AI literacy and benefit concerned the individual persona, which shaped the size of the observed gaps. Third, the dialogue corpus is uneven, with three policies accounting for more than 90\% of conversations and a few low-rating personas recurring across many pairings. Fourth, the dialogues' cooperative openings are consistent with LLMs' documented sycophancy~\cite{sharma2024towards} and tendency toward consensus~\cite{chuang2024simulating}, and the instruction to add new perspectives likely further encouraged agreement and elaboration, so the conversations may understate genuine disagreement. Finally, the LLooM concepts were generated automatically and served only as probes for interpretation.

Future work should compare persona-grounded evaluations with human responses to the same policies, including from adults with low digital skills, validate inferred opinion change against manually annotated dialogue turns, and test whether less cooperative dialogue settings produce genuine disagreement. Extending the design to other models (e.g., open source) and technical capacities (e.g., retrieval augmented generation on community town hall transcripts) and jurisdictions (e.g., provincial) would show whether the concept of conditional endorsement generalizes.

%TC:ignore
\bibliographystyle{ACM-Reference-Format}
\bibliography{sample-base}

\clearpage
\onecolumn
\appendix
\section{Self-Assessed Skill and Over-Confidence}\label{app:overconfidence}
This appendix reports a complementary analysis of the PIAAC Cycle~2 data that asks which Canadian adults rate their skill above what their measured proficiency supports. Critical evaluation of AI output is a core AI literacy competency~\cite{long2020ai}, self-efficacy is part of AI literacy in its own right~\cite{carolus2023mails}, and people who trust an automated system too readily tend to stop checking its output~\cite{parasuraman2010complacency}. PIAAC does not ask about AI, but it asks respondents to judge their performance on the assessment they have just completed, alongside a direct measure of that performance. We use over-confidence about one's own assessment performance as a proxy for the risk of over-trusting AI, not as a measure of it, and the analysis is descriptive throughout. The design adapts that of Khan and colleagues~\cite{khan2017understanding}, who studied the gap between subjective and objective financial knowledge among older Canadians with nested survey-weighted regressions and predicted profiles for recognizable groups. We keep the nested models and the profiles but do not use their difference score, which correlates mechanically with the objective measure through regression to the mean~\cite{gignac2020dunning}.
 
\subsection{Analytic Sample}\label{app:oc-data}
The analysis uses the Canadian sample of the PIAAC Cycle~2 public-use file~\cite{oecd2025technical}, 11,697 respondents in all. Rather than fixing one analytic sample up front, we start from the full sample and add exclusions only as each model requires them. Dropping respondents routed to the doorstep interview, who never saw the self-assessment question, and those never administered the post-assessment Effort module leaves 10,455 adults (Table~\ref{tab:oc-sample}). Model~1 drops those missing any of its covariates, leaving 9,883 (Table~\ref{tab:oc-ledger1}), and Model~2 drops respondents without work-context data or a scored occupation, leaving 6,825 (Table~\ref{tab:oc-ledger2}). The adults who never reached the self-assessment item sit well below the retained sample on proficiency (Table~\ref{tab:oc-lost}), so the results describe adults who reached the item, not the full population.
 
\begin{table}[htbp]
\centering\small
\caption{Derivation of the self-efficacy analytic sample.}\label{tab:oc-sample}
\begin{tabular}{lrr}
\toprule
Step & $n$ remaining & Excluded \\
\midrule
All records & 11,697 &  \\
Not a doorstep interview & 11,607 & 90 \\
Completed all three cognitive domains & 11,607 & 0 \\
Administered the Effort module & 10,455 & 1,152 \\
Answered the self-assessment question & 10,455 & 0 \\
\midrule
Analytic sample & 10,455 & 1,242 \\
\bottomrule
\end{tabular}
\end{table}
 
\subsection{Measures and Models}\label{app:oc-methods}
\paragraph{Estimation.}
Every quantity in this appendix is estimated separately on each of the ten plausible values and pooled by Rubin's rules, with standard errors from the 80 Fay-adjusted replicate weights (VEMETHODN~$=4$, VEFAYFAC~$=0.3$, $c \approx 0.02551$ for Canada in Cycle~2).
 
\paragraph{Self-assessment and proficiency.}
The background questionnaire asks each respondent to place the share of assessment items they answered correctly in one of five ordered bands (\texttt{EffortP001S}, from 20\% or less to 81--100\%). The public-use file carries no per-item correctness, so we examine \emph{relative calibration}: whether a respondent rates their performance higher than their measured proficiency would lead one to expect. Because the item refers to the assessment as a whole, the benchmark is a General Cognitive Proficiency Index (GCPI), the mean of the literacy, numeracy, and adaptive problem solving scores, built and standardized once per plausible value and never averaged into a single score.
 
\paragraph{Nested models.}
Rather than differencing the self-estimate against proficiency, we model the self-assessment band as a function of GCPI and covariates with an ordered (proportional-odds) logit. It is fit by maximum likelihood on each plausible value and each replicate weight, so standard errors come from the replicate spread and the between-plausible-value variance rather than the likelihood's Hessian; a survey-weighted ordered logit (\texttt{survey::svyolr}~\cite{lumley2004analysis}) reproduces the coefficients (Table~\ref{tab:oc-svyolr}). \textbf{Model~0} uses GCPI alone. \textbf{Model~1}, the primary specification, adds assessment effort (\texttt{EffortP002S}) as a calibration control, self-rated health, ICT use at home, the five BFI-2 personality traits, age, gender, education, and immigrant generation. \textbf{Model~2} restricts to employed adults with a scored occupation and adds occupational AI exposure, ICT use at work, task discretion, and learning at work. Coefficients are log-odds per standard deviation with two-sided Wald \emph{p}-values on the design-based standard errors, and no pair of continuous predictors is collinear (Table~\ref{tab:oc-vif}).
 
\paragraph{Occupational AI exposure.}
We use the AI Occupational Exposure index of Felten and colleagues~\cite{felten2021occupational}, which scores 774 US occupations of the 2010 Standard Occupational Classification (SOC) on how closely their required abilities relate to ten common AI applications. The Canadian public-use file reports occupation only as the ISCO-08 major group (\texttt{ISCO1C}), so each score is carried from SOC to ISCO-08 unit groups through the crosswalk of Hardy and colleagues~\cite{hardy2018educational} and averaged up to the nine major groups, with a SOC occupation that maps to several unit groups contributing an equal share to each. Scores carried through the US Bureau of Labor Statistics crosswalk~\cite{bls2015soc} differ by at most 0.005 on any major group. Armed forces occupations have no score, and the major-group score is standardized within the Model~2 sample.
 
\subsection{Self-Assessment and Measured Proficiency}\label{app:oc-results}
 
\begin{figure}[htbp]
\centering
\includegraphics[width=0.8\linewidth]{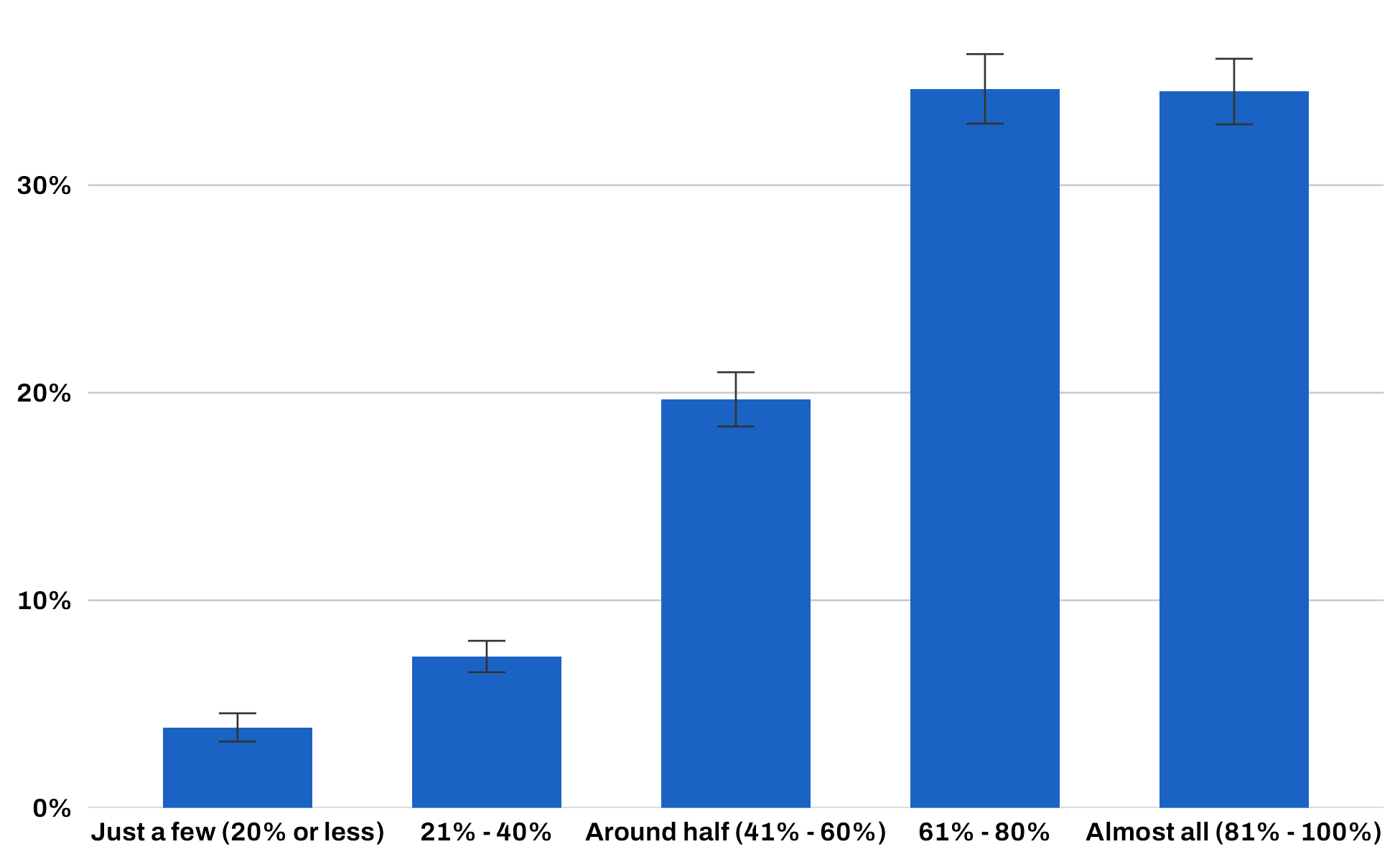}
\caption{Weighted distribution of self-assessed performance bands, with 95\% confidence intervals.}
\label{fig:oc-bands}
\end{figure}

\begin{figure}[htbp]
\centering
\includegraphics[width=0.8\linewidth]{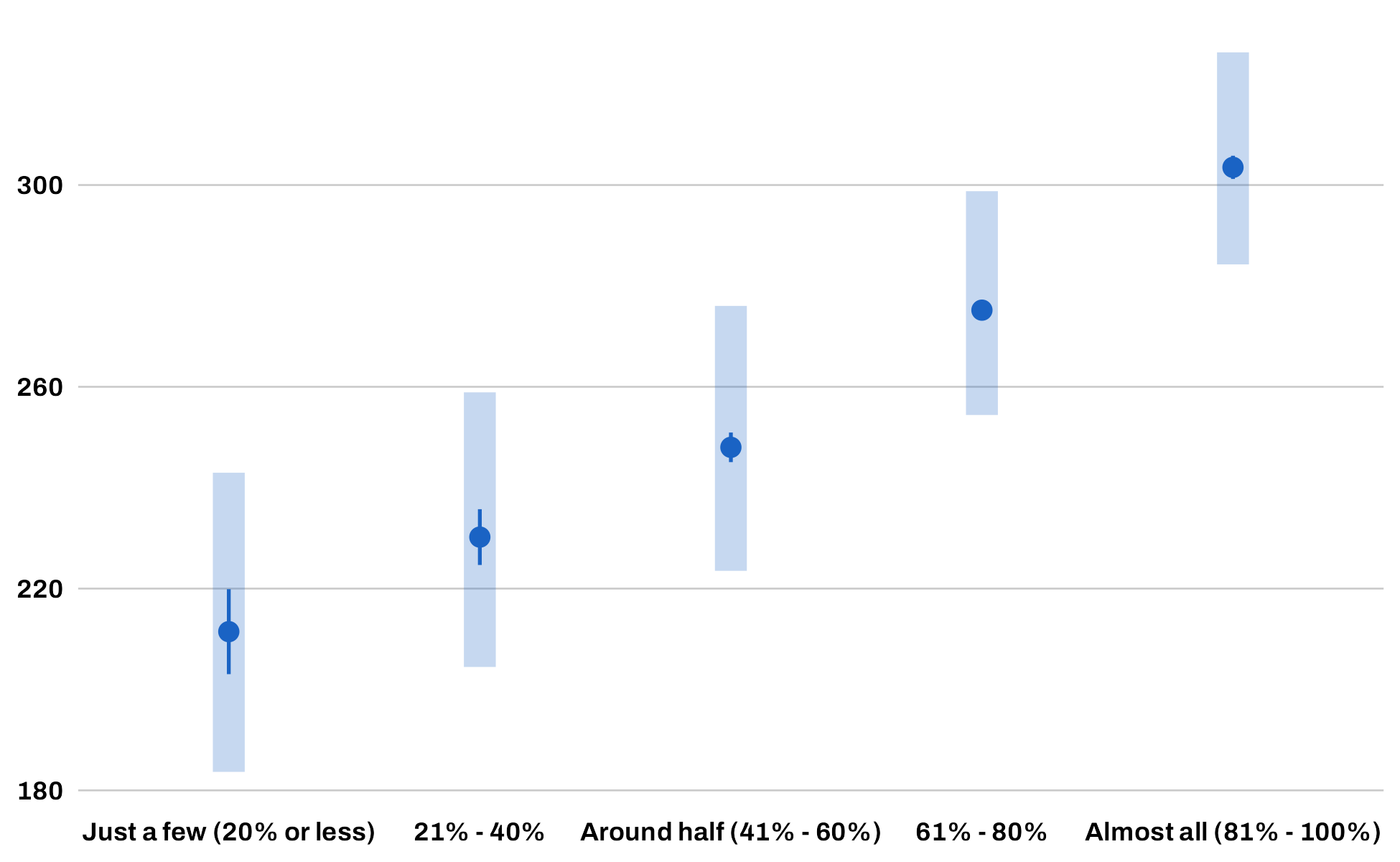}
\caption{Pooled GCPI proficiency by self-assessed performance band, with 95\% confidence intervals (points) and the 25th--75th percentile range (shaded).}
\label{fig:oc-calibration}
\end{figure}

Most adults rate themselves favorably: 69\% place themselves in the top two bands, claiming at least three-fifths of the assessment correct (Figure~\ref{fig:oc-bands}). Confidence tracks ability, with a one-standard-deviation increase in GCPI raising the log-odds of a higher band by 1.38 in Model~0 (Table~\ref{tab:oc-model0}), but the relationship is loose. Mean GCPI rises across the bands, yet adults of similar proficiency place themselves across several adjacent bands and the interquartile ranges overlap heavily (Figure~\ref{fig:oc-calibration}).
 
In Model~1, proficiency remains dominant (1.28 log-odds per SD; Table~\ref{tab:oc-model1}). The clearest association beyond proficiency is assessment effort (0.83), and women rate themselves well below men at equal proficiency ($-$0.76). ICT use at home, conscientiousness, openness, and emotional stability are near zero, and only extraversion reaches conventional significance, modestly. Net of proficiency, self-assessment moves with engagement and gender rather than digital exposure or personality. The GCPI slope is similar at each band threshold (Table~\ref{tab:oc-propodds}), and the fitted model reproduces the observed band distribution to within about a percentage point (Table~\ref{tab:oc-predobs}). Among employed adults, none of the work-context terms in Model~2 adds much (Table~\ref{tab:oc-model2}): occupational AI exposure is small and, if anything, negative ($-$0.095, SE 0.065), though exposure enters only at the major-group level.
 
\begin{table}[htbp]
\centering\small
\caption{Model 0 regression output, an ordered logit of self-assessment on GCPI alone (log-odds per SD).}\label{tab:oc-model0}
\begin{tabular}{lrrr}
\toprule
Term & Estimate & SE & \emph{p}-value \\
\midrule
GCPI & 1.383 & 0.057 & $<$0.001 \\
\bottomrule
\end{tabular}
\par\smallskip\footnotesize\raggedright \emph{p}-value: two-sided Wald test (estimate / SE) against the standard normal.
\end{table}
 
\begin{table}[htbp]
\centering\small
\caption{Model 1 regression output with all ordered-logit coefficients, in log-odds per SD for continuous predictors and versus the omitted reference for factors (age 35--44, upper secondary education, non-immigrant, male, excellent health).}\label{tab:oc-model1}
\begin{tabular}{lrrr}
\toprule
Term & Estimate & SE & \emph{p}-value \\
\midrule
GCPI & 1.278 & 0.062 & $<$0.001 \\
ICT use at home & 0.057 & 0.041 & 0.164 \\
Conscientiousness & 0.055 & 0.038 & 0.145 \\
Openness & 0.085 & 0.039 & 0.028 \\
Emotional stability & 0.064 & 0.041 & 0.116 \\
Extraversion & 0.125 & 0.041 & 0.003 \\
Agreeableness & $-$0.065 & 0.035 & 0.064 \\
Assessment effort & 0.831 & 0.049 & $<$0.001 \\
Health: very good & $-$0.141 & 0.117 & 0.229 \\
Health: good & $-$0.361 & 0.130 & 0.005 \\
Health: fair & $-$0.240 & 0.178 & 0.176 \\
Health: poor & $-$0.353 & 0.246 & 0.152 \\
Age 24 or less & 0.374 & 0.130 & 0.004 \\
Age 25--34 & $-$0.057 & 0.106 & 0.594 \\
Age 45--54 & $-$0.157 & 0.099 & 0.114 \\
Age 55+ & $-$0.102 & 0.121 & 0.403 \\
Female & $-$0.756 & 0.069 & $<$0.001 \\
Lower secondary or less & $-$0.086 & 0.145 & 0.554 \\
Post-secondary, non-tertiary & $-$0.038 & 0.099 & 0.701 \\
Tertiary, professional & 0.100 & 0.096 & 0.300 \\
Tertiary, bachelor & 0.492 & 0.126 & $<$0.001 \\
Tertiary, master/research & 0.531 & 0.152 & $<$0.001 \\
1st-generation immigrant & 0.064 & 0.104 & 0.540 \\
2nd-generation immigrant & $-$0.042 & 0.136 & 0.755 \\
One foreign-born parent & 0.125 & 0.153 & 0.414 \\
\bottomrule
\end{tabular}
\par\smallskip\footnotesize\raggedright \emph{p}-value: two-sided Wald test (estimate / SE) against the standard normal.
\end{table}
 
\subsection{Who Over-Rates Their Skill}\label{app:oc-who}
To identify who over-rates their skill, we do not use each adult's Model~1 residual, for two reasons. Because Model~1 adjusts for gender, age, and education, those characteristics shift each adult's benchmark rather than their measured over-confidence, so profiling the residual on them largely restates the model. The five-band scale also caps the residual, leaving the most room to overshoot to adults the model expects to rate themselves low. Instead, the benchmark is proficiency alone, modeled as an ordered logit of the self-assessment band on a natural spline of GCPI, and the comparison is made on the model's latent scale through the surrogate residual of Liu and Zhang~\cite{liu2018residuals}, which has no ceiling. An adult counts as over-confident when their residual falls in the top tenth of its distribution under the model; since a reported band pins the latent value only to an interval, each adult receives the exact probability of falling in that tail. Roughly one adult in ten is over-confident by design, and the share is flat across proficiency, between 9.8\% and 10.7\% in every GCPI quintile (Table~\ref{tab:oc-quintile}).
 
A weighted fractional logit~\cite{papke1996econometric} then relates each adult's over-confidence probability to proficiency and the Model~1 covariates (Table~\ref{tab:oc-personamodel}). Men are over-confident more often than women (12.1\% against 8.3\%; Table~\ref{tab:oc-group}), and education raises over-confidence even with proficiency held fixed: a bachelor's degree adds 0.40 log-odds while each standard deviation of proficiency subtracts 0.22. Adults under 25 over-rate more than those aged 35 to 44 (0.38) once other characteristics are held fixed. Assessment effort is the strongest predictor, from 4.3\% among adults reporting the least effort to 15.6\% among those reporting the most, which likely reflects respondents reading their effort as success, so we treat it as a control rather than a target.
 
Following Khan and colleagues~\cite{khan2017understanding}, we fix the characteristics a program can target (age, gender, and education) and average the predicted share over everything else (Table~\ref{tab:oc-personas}). Men with a bachelor's degree are over-confident at about 14.6\% whether aged 25 to 34 or 55 and over, as are men under 25 with upper secondary schooling, while women aged 55 and over with lower secondary schooling or less are the least likely, at 5.3\%. This reverses the education gradient Khan and colleagues found for financial knowledge, where the over-confident profile was the older man with a high school education or less, so groups to target for AI calibration cannot be read off from financial literacy work.
 
\begin{table}[htbp]
\centering\small
\caption{Predicted share over-confident relative to measured skill for persona profiles (\%), averaged over the observed distribution of every other characteristic.}\label{tab:oc-personas}
\begin{tabular}{lrr}
\toprule
Persona & Estimate & SE \\
\midrule
All adults & 10.2 & 0.4 \\
Men 25--34, bachelor's degree & 14.6 & 1.3 \\
Men 55 and over, bachelor's degree & 14.5 & 1.3 \\
Men 24 or less, upper secondary & 14.5 & 1.7 \\
Women 25--34, bachelor's degree & 9.7 & 0.8 \\
Women 24 or less, upper secondary & 9.6 & 1.2 \\
Women 55 and over, lower secondary or less & 5.3 & 0.7 \\
\bottomrule
\end{tabular}
\end{table}
 
\subsection{Robustness and Limitations}\label{app:oc-limitations}
The Model~1 proficiency coefficient is similar whether the benchmark is GCPI or literacy, numeracy, or adaptive problem solving alone (Table~\ref{tab:oc-domain}), and fitting Model~1 as a linear regression of the band score reproduces the ordered logit's signs and significance (Table~\ref{tab:oc-linear}).
 
Four limitations apply. First, plausible values are generated for proficiency as an outcome, and using them as a predictor attenuates the proficiency--confidence slope, most for the roughly one-third of the sample whose score in a given domain is imputed (Table~\ref{tab:oc-coverage}). Second, occupational AI exposure is available only at the ISCO major group, whose occupations span as much as 4.2 standard deviations of exposure, and the crosswalk assumes that US exposure carries over to Canadian jobs. Third, PIAAC measures neither AI use nor verification behavior, so the link from miscalibration to AI over-trust is a theoretically motivated proxy. Fourth, the self-assessment item was not reached by doorstep respondents, non-completers, and adults routed through the easier adaptive paths, so the sample is positively selected on ability (Table~\ref{tab:oc-lost}).
 
\subsection{Supplementary Tables}\label{app:oc-supplementary}
\begin{table}[htbp]
\centering\small
\caption{Model 1 exclusion ledger, with covariate completeness layered on the analytic sample.}\label{tab:oc-ledger1}
\begin{tabular}{lrr}
\toprule
Step & $n$ remaining & Excluded \\
\midrule
All records & 10,455 &  \\
ICT use at home reported & 10,356 & 99 \\
Conscientiousness scale derived & 10,322 & 34 \\
Openness scale derived & 10,222 & 100 \\
Emotional stability scale derived & 10,210 & 12 \\
Extraversion scale derived & 10,196 & 14 \\
Agreeableness scale derived & 10,174 & 22 \\
Self-rated health recorded & 10,171 & 3 \\
Immigrant generation recorded & 10,115 & 56 \\
Engagement with the assessment reported & 9,883 & 232 \\
Age band recorded & 9,883 & 0 \\
Gender recorded & 9,883 & 0 \\
Education recorded (EDCAT6\_TC1) & 9,883 & 0 \\
\midrule
Analytic sample & 9,883 & 572 \\
\bottomrule
\end{tabular}
\end{table}
 
\begin{table}[htbp]
\centering\small
\caption{Model 2 exclusion ledger, layered on Model 1's retained sample so each row shows only what the new rule drops.}\label{tab:oc-ledger2}
\begin{tabular}{lrr}
\toprule
Step & $n$ remaining & Excluded \\
\midrule
All records & 9,883 &  \\
ICT use at work reported & 7,458 & 2,425 \\
Occupation in a scored ISCO major group & 6,840 & 618 \\
Task discretion reported & 6,838 & 2 \\
Learning at work reported & 6,825 & 13 \\
\midrule
Analytic sample & 6,825 & 3,058 \\
\bottomrule
\end{tabular}
\end{table}
 
\begin{table}[htbp]
\centering\small
\caption{Model 2 regression output, adding the work-context terms to Model 1 on the employed subsample (log-odds per SD). Factor references as in Model 1.}\label{tab:oc-model2}
\begin{tabular}{lrrr}
\toprule
Term & Estimate & SE & \emph{p}-value \\
\midrule
GCPI & 1.352 & 0.074 & $<$0.001 \\
ICT use at home & 0.028 & 0.054 & 0.602 \\
Conscientiousness & 0.032 & 0.053 & 0.546 \\
Openness & 0.030 & 0.051 & 0.551 \\
Emotional stability & 0.108 & 0.059 & 0.064 \\
Extraversion & 0.139 & 0.052 & 0.007 \\
Agreeableness & $-$0.053 & 0.048 & 0.267 \\
Assessment effort & 0.866 & 0.059 & $<$0.001 \\
Health: very good & $-$0.056 & 0.133 & 0.677 \\
Health: good & $-$0.202 & 0.158 & 0.201 \\
Health: fair & $-$0.158 & 0.206 & 0.442 \\
Health: poor & 0.033 & 0.309 & 0.916 \\
Age 24 or less & 0.245 & 0.161 & 0.129 \\
Age 25--34 & $-$0.102 & 0.124 & 0.408 \\
Age 45--54 & $-$0.202 & 0.109 & 0.064 \\
Age 55+ & $-$0.139 & 0.160 & 0.386 \\
Female & $-$0.721 & 0.074 & $<$0.001 \\
Lower secondary or less & $-$0.219 & 0.205 & 0.287 \\
Post-secondary, non-tertiary & $-$0.043 & 0.141 & 0.762 \\
Tertiary, professional & 0.020 & 0.118 & 0.863 \\
Tertiary, bachelor & 0.523 & 0.147 & $<$0.001 \\
Tertiary, master/research & 0.521 & 0.166 & 0.002 \\
1st-generation immigrant & 0.027 & 0.129 & 0.832 \\
2nd-generation immigrant & $-$0.064 & 0.175 & 0.716 \\
One foreign-born parent & 0.157 & 0.176 & 0.370 \\
ICT use at work & 0.056 & 0.063 & 0.374 \\
Occupational AI exposure & $-$0.095 & 0.065 & 0.142 \\
Task discretion & $-$0.007 & 0.037 & 0.847 \\
Learning at work & $-$0.020 & 0.061 & 0.741 \\
\bottomrule
\end{tabular}
\par\smallskip\footnotesize\raggedright \emph{p}-value: two-sided Wald test (estimate / SE) against the standard normal.
\end{table}
 
\begin{table}[htbp]
\centering\small
\caption{Share of adults over-confident relative to their measured skill, by GCPI quintile (\%).}\label{tab:oc-quintile}
\begin{tabular}{lrr}
\toprule
GCPI quintile & Estimate & SE \\
\midrule
Quintile 1 & 10.6 & 1.4 \\
Quintile 2 & 10.7 & 1.3 \\
Quintile 3 & 9.8 & 0.8 \\
Quintile 4 & 10.1 & 0.5 \\
Quintile 5 & 9.9 & 0.3 \\
\bottomrule
\end{tabular}
\end{table}
 
\begin{table}[htbp]
\centering\small
\caption{Share of adults over-confident relative to their measured skill, by characteristic (\%).}\label{tab:oc-group}
\begin{tabular}{llrr}
\toprule
Characteristic & Group & Estimate & SE \\
\midrule
Gender & Men & 12.1 & 0.6 \\
 & Women & 8.3 & 0.4 \\
\addlinespace
Age & 24 or less & 9.9 & 0.9 \\
 & 25--34 & 9.3 & 0.7 \\
 & 35--44 & 10.7 & 0.7 \\
 & 45--54 & 10.3 & 1.2 \\
 & 55 and over & 10.8 & 0.7 \\
\addlinespace
Education & Lower secondary or less & 8.1 & 1.0 \\
 & Upper secondary & 9.5 & 0.9 \\
 & Post-secondary, non-tertiary & 9.2 & 0.7 \\
 & Tertiary, professional & 11.0 & 0.8 \\
 & Bachelor's & 11.6 & 0.8 \\
 & Master's or research & 11.6 & 1.0 \\
\addlinespace
Immigrant generation & Non-immigrant & 9.5 & 0.5 \\
 & One foreign-born parent & 10.2 & 1.0 \\
 & 2nd-generation immigrant & 8.7 & 1.1 \\
 & 1st-generation immigrant & 12.1 & 0.8 \\
\addlinespace
Self-rated health & Excellent & 12.2 & 0.9 \\
 & Very good & 10.7 & 0.7 \\
 & Good & 8.8 & 0.6 \\
 & Fair & 10.4 & 1.3 \\
 & Poor & 5.7 & 1.3 \\
\addlinespace
Assessment effort (1--10) & 1--6 & 4.3 & 0.5 \\
 & 7--8 & 7.8 & 0.4 \\
 & 9--10 & 15.6 & 0.8 \\
\bottomrule
\end{tabular}
\end{table}
 
\begin{table}[htbp]
\centering\small
\caption{Persona model, a fractional logit of the over-confidence probability on measured proficiency and the Model 1 covariates, in log-odds per SD for continuous predictors and versus the Model 1 reference levels for factors.}\label{tab:oc-personamodel}
\begin{tabular}{lrrr}
\toprule
Term & Estimate & SE & \emph{p}-value \\
\midrule
GCPI & $-$0.224 & 0.052 & $<$0.001 \\
ICT use at home & 0.026 & 0.041 & 0.522 \\
Conscientiousness & 0.060 & 0.038 & 0.113 \\
Openness & 0.003 & 0.034 & 0.933 \\
Emotional stability & 0.005 & 0.047 & 0.909 \\
Extraversion & 0.047 & 0.040 & 0.249 \\
Agreeableness & $-$0.026 & 0.039 & 0.500 \\
Assessment effort & 0.622 & 0.053 & $<$0.001 \\
Health: very good & $-$0.128 & 0.111 & 0.246 \\
Health: good & $-$0.315 & 0.123 & 0.011 \\
Health: fair & $-$0.046 & 0.155 & 0.767 \\
Health: poor & $-$0.463 & 0.262 & 0.076 \\
Age 24 or less & 0.382 & 0.131 & 0.003 \\
Age 25--34 & $-$0.005 & 0.111 & 0.962 \\
Age 45--54 & $-$0.062 & 0.120 & 0.605 \\
Age 55+ & $-$0.018 & 0.104 & 0.862 \\
Female & $-$0.481 & 0.081 & $<$0.001 \\
Lower secondary or less & $-$0.269 & 0.154 & 0.080 \\
Post-secondary, non-tertiary & 0.033 & 0.146 & 0.819 \\
Tertiary, professional & 0.184 & 0.135 & 0.172 \\
Tertiary, bachelor & 0.398 & 0.131 & 0.002 \\
Tertiary, master/research & 0.388 & 0.133 & 0.004 \\
1st-generation immigrant & 0.083 & 0.090 & 0.353 \\
2nd-generation immigrant & $-$0.064 & 0.128 & 0.618 \\
One foreign-born parent & 0.053 & 0.142 & 0.706 \\
\bottomrule
\end{tabular}
\par\smallskip\footnotesize\raggedright \emph{p}-value: two-sided Wald test (estimate / SE) against the standard normal.
\end{table}
 
\begin{table}[htbp]
\centering\small
\caption{The Model 1 proficiency coefficient under each benchmark, the composite GCPI versus each domain alone in its place.}\label{tab:oc-domain}
\begin{tabular}{lrr}
\toprule
Proficiency benchmark & Estimate & SE \\
\midrule
Composite (GCPI) & 1.278 & 0.062 \\
Literacy & 1.107 & 0.062 \\
Numeracy & 1.198 & 0.056 \\
Adaptive problem solving & 1.108 & 0.054 \\
\bottomrule
\end{tabular}
\end{table}
 
\begin{table}[htbp]
\centering\small
\caption{Measured proficiency of respondents dropped by Effort-module non-administration versus those retained. The sample is positively selected on ability.}\label{tab:oc-lost}
\begin{tabular}{lrr}
\toprule
Group & Estimate & SE \\
\midrule
Dropped & 233.3 & 3.0 \\
Retained & 273.9 & 0.7 \\
\bottomrule
\end{tabular}
\end{table}
 
\begin{table}[htbp]
\centering\small
\caption{Share of the analytic sample administered each cognitive domain directly, rather than by imputation from the background questionnaire.}\label{tab:oc-coverage}
\begin{tabular}{lrr}
\toprule
Domain & Estimate & SE \\
\midrule
Literacy & 0.654 & 0.010 \\
Numeracy & 0.678 & 0.008 \\
Adaptive problem solving & 0.652 & 0.009 \\
\bottomrule
\end{tabular}
\end{table}
 
\begin{table}[htbp]
\centering\small
\caption{Variance-inflation factors and the largest absolute weighted correlation for each continuous predictor. No pair is collinear.}\label{tab:oc-vif}
\begin{tabular}{lrlr}
\toprule
Predictor & VIF & Most correlated with & $|r|$ \\
\midrule
GCPI & 1.12 & Assessment effort & 0.19 \\
ICT use at home & 1.06 & Extraversion & 0.15 \\
Conscientiousness & 1.35 & Emotional stability & 0.35 \\
Openness & 1.08 & Extraversion & 0.17 \\
Emotional stability & 1.26 & Conscientiousness & 0.35 \\
Extraversion & 1.24 & Conscientiousness & 0.34 \\
Agreeableness & 1.20 & Conscientiousness & 0.33 \\
Assessment effort & 1.07 & GCPI & 0.19 \\
\bottomrule
\end{tabular}
\end{table}
 
\begin{table}[htbp]
\centering\small
\caption{Cross-check of the custom replicate engine against survey::svyolr on one plausible value. Point estimates match and standard errors agree closely.}\label{tab:oc-svyolr}
\begin{tabular}{lrrrr}
\toprule
Term & Custom est. & Custom SE & svyolr est. & svyolr SE \\
\midrule
GCPI & 1.271 & 0.065 & 1.271 & 0.063 \\
ICT use at home & 0.053 & 0.039 & 0.053 & 0.039 \\
Conscientiousness & 0.052 & 0.037 & 0.052 & 0.037 \\
Openness & 0.079 & 0.038 & 0.079 & 0.038 \\
Emotional stability & 0.065 & 0.040 & 0.065 & 0.039 \\
Extraversion & 0.133 & 0.041 & 0.133 & 0.041 \\
Agreeableness & $-$0.070 & 0.034 & $-$0.070 & 0.034 \\
Assessment effort & 0.830 & 0.051 & 0.830 & 0.050 \\
Health: very good & $-$0.138 & 0.119 & $-$0.138 & 0.119 \\
Health: good & $-$0.353 & 0.129 & $-$0.353 & 0.128 \\
Health: fair & $-$0.212 & 0.174 & $-$0.212 & 0.174 \\
Health: poor & $-$0.324 & 0.261 & $-$0.324 & 0.259 \\
Age 24 or less & 0.383 & 0.125 & 0.383 & 0.119 \\
Age 25--34 & $-$0.059 & 0.106 & $-$0.059 & 0.105 \\
Age 45--54 & $-$0.162 & 0.094 & $-$0.162 & 0.093 \\
Age 55+ & $-$0.089 & 0.118 & $-$0.089 & 0.117 \\
Female & $-$0.752 & 0.072 & $-$0.752 & 0.071 \\
Lower secondary or less & $-$0.112 & 0.140 & $-$0.112 & 0.140 \\
Post-secondary, non-tertiary & $-$0.056 & 0.101 & $-$0.056 & 0.099 \\
Tertiary, professional & 0.088 & 0.102 & 0.088 & 0.102 \\
Tertiary, bachelor & 0.499 & 0.128 & 0.499 & 0.128 \\
Tertiary, master/research & 0.536 & 0.151 & 0.536 & 0.151 \\
1st-generation immigrant & 0.059 & 0.109 & 0.059 & 0.108 \\
2nd-generation immigrant & $-$0.044 & 0.133 & $-$0.044 & 0.132 \\
One foreign-born parent & 0.107 & 0.154 & 0.107 & 0.154 \\
\bottomrule
\end{tabular}
\end{table}
 
\begin{table}[htbp]
\centering\small
\caption{The GCPI slope estimated separately at each band threshold, a check on the proportional-odds assumption. The slopes stay close, so the single ordered-logit summary is fair.}\label{tab:oc-propodds}
\begin{tabular}{lrr}
\toprule
Threshold & Estimate & SE \\
\midrule
Band $\geq$ 2 & 1.249 & 0.091 \\
Band $\geq$ 3 & 1.293 & 0.069 \\
Band $\geq$ 4 & 1.400 & 0.066 \\
Band $\geq$ 5 & 1.470 & 0.081 \\
\bottomrule
\end{tabular}
\end{table}
 
\begin{table}[htbp]
\centering\small
\caption{Predicted versus observed weighted share (\%) in each self-assessment band, Model 1.}\label{tab:oc-predobs}
\begin{tabular}{lrrrr}
\toprule
Band & Observed & SE & Predicted & SE \\
\midrule
1 & 3.9 & 0.3 & 3.4 & 0.3 \\
2 & 7.3 & 0.4 & 6.7 & 0.4 \\
3 & 19.7 & 0.7 & 18.8 & 0.7 \\
4 & 34.6 & 0.9 & 35.5 & 0.9 \\
5 & 34.5 & 0.8 & 35.6 & 0.9 \\
\bottomrule
\end{tabular}
\end{table}
 
\begin{table}[htbp]
\centering\small
\caption{Model 1 as an ordered logit (log-odds) and as a linear regression of the band score (bands), side by side.}\label{tab:oc-linear}
\begin{tabular}{lrrrr}
\toprule
 & \multicolumn{2}{c}{Ordered logit} & \multicolumn{2}{c}{Linear} \\
\cmidrule(lr){2-3}\cmidrule(lr){4-5}
Term & Estimate & SE & Estimate & SE \\
\midrule
ICT use at home & 0.057 & 0.041 & 0.026 & 0.016 \\
Conscientiousness & 0.055 & 0.038 & 0.021 & 0.015 \\
Openness & 0.085 & 0.039 & 0.028 & 0.015 \\
Emotional stability & 0.064 & 0.041 & 0.012 & 0.015 \\
Extraversion & 0.125 & 0.041 & 0.052 & 0.016 \\
Agreeableness & $-$0.065 & 0.035 & $-$0.025 & 0.013 \\
Assessment effort & 0.831 & 0.049 & 0.344 & 0.021 \\
Health: very good & $-$0.141 & 0.117 & $-$0.037 & 0.040 \\
Health: good & $-$0.361 & 0.130 & $-$0.116 & 0.050 \\
Health: fair & $-$0.240 & 0.178 & $-$0.085 & 0.063 \\
Health: poor & $-$0.353 & 0.246 & $-$0.177 & 0.107 \\
Age 24 or less & 0.374 & 0.130 & 0.177 & 0.050 \\
Age 25--34 & $-$0.057 & 0.106 & $-$0.008 & 0.038 \\
Age 45--54 & $-$0.157 & 0.099 & $-$0.054 & 0.039 \\
Age 55+ & $-$0.102 & 0.121 & $-$0.052 & 0.046 \\
Female & $-$0.756 & 0.069 & $-$0.283 & 0.025 \\
Lower secondary or less & $-$0.086 & 0.145 & $-$0.055 & 0.061 \\
Post-secondary, non-tertiary & $-$0.038 & 0.099 & 0.009 & 0.038 \\
Tertiary, professional & 0.100 & 0.096 & 0.070 & 0.037 \\
Tertiary, bachelor & 0.492 & 0.126 & 0.175 & 0.046 \\
Tertiary, master/research & 0.531 & 0.152 & 0.204 & 0.053 \\
1st-generation immigrant & 0.064 & 0.104 & 0.010 & 0.037 \\
2nd-generation immigrant & $-$0.042 & 0.136 & $-$0.006 & 0.053 \\
One foreign-born parent & 0.125 & 0.153 & 0.053 & 0.055 \\
\bottomrule
\end{tabular}
\end{table}

\clearpage

\section{Methodological Details}\label{app:methods}

\subsection{Identifying Policies and Landscape Challenges}\label{app:policies}
We identified passages in \emph{Canada's National Artificial Intelligence Strategy: AI for All} that described government actions, including regulatory reforms, investments, programs, institutional changes, partnerships, and measurable targets. We treated a passage as a policy when the document specified an action that the government had taken, had committed to take, or planned to pursue; general statements about AI's potential benefits were not coded as policies. We preserved distinctions among existing initiatives, newly announced measures, and future commitments, and recorded their page locations. We then iteratively consolidated overlapping actions into broader policy areas, while retaining named programs as separate items where appropriate.

We also analyzed the document for references to risks, harms, barriers, gaps, dependencies, vulnerabilities, and resource constraints. We grouped these challenges into recurring areas: public trust and AI literacy, employment disruption, privacy and online harms, unequal access, business adoption, foreign technological dependence, infrastructure and environmental constraints, fragmented data, and difficulties in retaining talent and commercializing Canadian AI research. We used an LLM to paraphrase the 20 policies and nine challenges into self-contained descriptions and stored them in separate indexed corpora. Tables~\ref{tab:policylist} and~\ref{tab:challengelist} list the resulting English descriptions; we also prepared French translations of both.

\begin{small}
\begin{longtable}{>{\raggedright\arraybackslash}p{0.27\linewidth}p{0.66\linewidth}}
\caption{The 20 policies identified in \emph{AI for All}.}\label{tab:policylist}\\
\toprule
Policy & Description as presented to agents \\
\midrule
\endfirsthead
\toprule
Policy & Description as presented to agents \\
\midrule
\endhead
\bottomrule
\endlastfoot
Consumer privacy reform & The Government of Canada announces plans to modernize consumer privacy legislation. The proposed changes would recognize privacy as a fundamental right, strengthen people's control over their personal information, and protect children's information from exploitation. The government also plans to address inappropriate uses of personal data, including surveillance pricing. \\\addlinespace[3pt]
Online safety legislation & The Government of Canada announces plans to introduce online safety laws addressing harms associated with digital technologies and AI. Its stated priorities include protecting children, providing legal tools to address deepfakes, making chatbot interactions safer, and holding those responsible for online harms accountable. It also commits to protecting elections against AI-enabled misinformation and foreign interference. \\\addlinespace[3pt]
AI safety evaluation and certification & The Government of Canada announces plans to create a Canada Trusted AI Certification program intended to help people identify trustworthy AI products. It also commits CAD 50 million to expand the Canadian AI Safety Institute's work on emerging risks, technical research, and transparent evaluations of AI models. \\\addlinespace[3pt]
Identifying AI systems and generated content & The Government of Canada announces that it will work on measures to make AI use more transparent, including capabilities such as watermarking AI-generated content. The stated aim is to help people recognize when they are interacting with an AI system or encountering AI-generated material. \\\addlinespace[3pt]
Free introductory AI education & The Government of Canada announces plans for a National AI Literacy Initiative offering free introductory AI education accessible to all Canadians. Public libraries and community organizations would help deliver learning opportunities, including in rural, remote, and northern communities. The initiative would cover practical AI use as well as recognizing bias, misinformation, privacy risks, and unsafe uses. \\\addlinespace[3pt]
AI tools for postsecondary students & The Government of Canada announces a commitment to give all postsecondary students access to trusted AI agents. It also plans to bring AI literacy content to one million students entering postsecondary education and provide classroom AI learning resources and training to more than 3,000 educators. \\\addlinespace[3pt]
AI-related employment and training & The Government of Canada announces a plan to support up to 90,000 AI-related employment and work-placement opportunities by 2031 through student employment, skills, and Mitacs programs. It also plans to assess and expand employer-led AI training for workers in the middle of their careers, including people in skilled trades, working with employers, unions, colleges, and other training providers. \\\addlinespace[3pt]
Indigenous leadership in AI & The Government of Canada announces plans to support Indigenous-led AI initiatives that strengthen Indigenous languages and cultural expression. It commits to working with Indigenous communities on their ability to shape how AI is developed and used in Indigenous contexts, including applications involving language, land management, and cultural heritage. \\\addlinespace[3pt]
AI support for Canadian creators & The Government of Canada announces plans to establish a CAD 50 million Creative Technology Program. The program is intended to support Canadian creators in using AI on their own terms, as part of a wider commitment to represent Canadian voices, languages, and cultural knowledge in AI development and use. \\\addlinespace[3pt]
English and French in government AI & The Government of Canada announces a commitment that AI systems deployed within government will perform equally well in English and French. It also plans to support the adoption of AI tools that protect and promote French, including its use in Quebec and Francophone minority communities elsewhere in Canada. \\\addlinespace[3pt]
Accessibility and inclusion in AI & The Government of Canada announces plans to promote a national standard on accessible and equitable AI and to address accessibility barriers in AI systems. It also commits to applying Gender-Based Analysis Plus across AI policy, skills development, innovation, and governance, including attention to groups that experience disproportionate harms or barriers. \\\addlinespace[3pt]
Support for small-business AI adoption & The Government of Canada announces plans to use the Business Development Bank of Canada's CAD 500 million LIFT program to help small and medium-sized businesses finance AI adoption. It also commits CAD 500 million to expand the Regional Artificial Intelligence Initiative, alongside tools and advice to help businesses identify uses for AI. The government sets a target for 60 percent of Canadian businesses to have adopted AI by 2034. \\\addlinespace[3pt]
AI mission focused on healthcare & The Government of Canada announces plans for an AI Missions Program, with CAD 200 million committed to its first mission on improving health outcomes. The mission would bring together researchers, Canadian AI companies, governments, and healthcare practitioners. The government's stated goals include improving access to primary care, reducing emergency-room waits, preventing avoidable visits, and reducing administrative work for physicians. \\\addlinespace[3pt]
Health-data platforms for research and AI & The Government of Canada announces plans to invest CAD 100 million in a Health Sector Data Space with the Canadian Institute for Health Information, linking secure, standardized health datasets for research and performance measurement. It also commits CAD 100 million to expand VITAL into five additional provinces, supporting the use of hospital clinical data for health research and AI development. The strategy commits to strong privacy protections for this data infrastructure. \\\addlinespace[3pt]
AI adoption in federal public services & The Government of Canada announces plans to accelerate the procurement and deployment of AI in federal public services through the Office of Digital Transformation. Its stated aims include faster application processing and reduced administrative work. The government commits to transparency, privacy, and accountability standards, with human oversight of meaningful decisions and tasks, and plans a Prime Minister's Innovation Fellows Program to recruit technical expertise. \\\addlinespace[3pt]
Canadian-controlled AI infrastructure & The Government of Canada announces plans to expand Canadian-controlled AI infrastructure, including a public supercomputer by 2031, domestic cloud services, and data centres. It will prioritize building key capabilities in Canada, while also partnering with trusted allies or purchasing existing solutions where appropriate. The strategy connects new data-centre development with clean-energy expansion, environmental standards, and benefits for local communities. \\\addlinespace[3pt]
AI research and talent recruitment & The Government of Canada announces plans to expand the Canada CIFAR AI Chairs program from 130 to nearly 200 researchers and strengthen Canada's national AI institutes. It also plans to expand the Global Talent Stream to accelerate the arrival and onboarding of highly skilled AI workers and align permanent-residency measures with retaining recruited talent. \\\addlinespace[3pt]
Financing and commercialization for Canadian AI companies & The Government of Canada announces plans for a CAD 500 million Canadian Tech Growth Fund to support the expansion of Canadian AI companies, including possible government equity investments. It also commits an additional CAD 700 million through the Compute Access Fund and CAD 130 million for commercialization programs at national AI institutes. Federal purchasing under the Buy Canadian policy would help provide customers for domestic firms. \\\addlinespace[3pt]
International cooperation on sovereign AI & The Government of Canada announces plans to expand the Sovereign Technology Alliance launched with Germany. The alliance would coordinate work on AI models, digital infrastructure, research, safety evaluations, and procurement opportunities among participating countries. The government describes the alliance as a way to increase technological choice and reduce dependence on dominant technology ecosystems. \\\addlinespace[3pt]
Support for open-source AI & The Government of Canada announces plans to work with other countries, researchers, civil society, industry, and AI institutes to sustain open-source AI development in the public interest. It also plans an inventory and shared library of AI tools to support adoption by researchers, small and medium-sized businesses, nonprofit organizations, and public-interest innovators. \\
\end{longtable}
\end{small}
 
\begin{small}
\begin{longtable}{>{\raggedright\arraybackslash}p{0.27\linewidth}p{0.66\linewidth}}
\caption{The nine landscape challenges identified in \emph{AI for All}.}\label{tab:challengelist}\\
\toprule
Challenge & Description as presented to agents \\
\midrule
\endfirsthead
\toprule
Challenge & Description as presented to agents \\
\midrule
\endhead
\bottomrule
\endlastfoot
Low public trust and AI literacy & Many Canadians have limited AI understanding, training, or confidence. The strategy reports that 24\% of Canadians have received AI training. \\\addlinespace[3pt]
Employment disruption & Workers may need to reskill as jobs change, entry-level roles contract, and some female-dominated sectors face particular exposure. \\\addlinespace[3pt]
Privacy, discrimination, and online harms & Risks include misuse of personal information, algorithmic discrimination, deepfake sexual violence, fraud, misinformation, and threats to democratic institutions. \\\addlinespace[3pt]
Unequal access and disproportionate harms & Indigenous peoples, racialized communities, and people with disabilities may face distinct barriers or harms, alongside cultural and linguistic representation concerns. \\\addlinespace[3pt]
Business adoption barriers & Businesses face costs, limited expertise, and uncertainty about useful applications. The strategy reports that 78\% of non-adopting firms do not see how AI benefits their goods or services. \\\addlinespace[3pt]
Dependence on foreign technology & Reliance on foreign cloud providers, chips, infrastructure, and models raises concerns about control, foreign legal regimes, vendor lock-in, and technological choice. \\\addlinespace[3pt]
Infrastructure and environmental constraints & AI expansion requires substantial computing, electricity, and potentially water, while available power can constrain infrastructure growth. \\\addlinespace[3pt]
Fragmented and inaccessible data & Useful datasets remain isolated across institutions and require common standards, secure infrastructure, and privacy protections. \\\addlinespace[3pt]
Talent retention and commercialization & Canada faces competition for researchers, limited growth capital, difficulties translating research into products, and relocation of firms abroad. \\
\end{longtable}
\end{small}

\subsection{PIAAC Data Preparation}\label{app:piaac}
We combined the raw PIAAC Cycle 2 public-use survey data with the international codebook to produce a readable Excel dataset. We mapped categorical response codes to their corresponding labels and did not impute missing-value codes. Using the final full-sample weight, we then drew a weighted sample of 3,200 records.

\subsection{Persona Construction}\label{app:personas}

Each persona drew on the sampled respondent's demographic background, language, education, employment, occupation, household and economic context, digital technology use, learning and training experiences, personality, social trust, health, and calibration measures. We converted categorical variables into labeled descriptions. For some numeric variables, such as years of formal education (\texttt{YRSQUALC2}), we retained exact values. We converted behavioral, workplace, and personality measures into lower, typical, or higher categories using the weighted sample's 25th and 75th percentiles. These measures included digital technology use at home (\texttt{ICTHOMEC2}), reading and numeracy at home (\texttt{READHOMEC2\_T1}, \texttt{NUMHOMEC2}), learning at work (\texttt{LEARNATWORKC2}), and openness (\texttt{OPEM}, \texttt{OPEM\_CURI}). We organized the variables into thematic sections, namely demography, education, employment, digital skills, learning and training, personality, and social and health characteristics. Each persona received a stable identifier and was stored in a JSON file (Figure~\ref{fig:persona}).

\begin{figure}[h]
\centering
\fbox{\begin{minipage}{0.95\linewidth}\footnotesize
\textbf{PIAAC-1754}\par\smallskip
\textbf{Demography:} age: 55 plus; gender: Male; immigration background: Non 1st or 2nd generation immigrants; questionnaire language: ENG; language alignment: Test language same as native language.\par
\textbf{Education:} educational attainment: ISCED 2.\par
\textbf{Employment:} employment status: Out of the labour force; occupation: No paid work for past 5 years.\par
\textbf{Household and economic context:} household: not living with spouse or partner.\par
\textbf{Core skills:} reading use at home: lower relative to the sample; numeracy use at home: lower relative to the sample.\par
\textbf{Digital skills:} ever used a digital device: No.\par
\textbf{Learning and training:} formal education in the last 12 months: Did not participate in FE; non-formal education in the last 12 months: Did not participate in NFE.\par
\textbf{Personality:} agreeableness: lower relative to the sample; conscientiousness: lower relative to the sample; emotional stability: lower relative to the sample; extraversion: lower relative to the sample; open-mindedness: lower relative to the sample; intellectual curiosity: lower relative to the sample; anxiety (weak general-affect context): typical relative to the sample.\par
\textbf{Social and health:} social trust (0--10): You can't be too careful; health: Good.\par
\end{minipage}}
\caption{Example persona constructed from PIAAC responses (PIAAC-1754, quoted in Section~\ref{sec:findings-skills}). Fields absent from a respondent's record, such as income, are omitted.}
\label{fig:persona}
\end{figure}

We assigned each persona a randomly selected subset of the nine landscape challenges (Figure~\ref{fig:challenges}): low public trust and AI literacy; employment disruption; privacy, discrimination, and online harms; unequal access and disproportionate harms; business adoption barriers; dependence on foreign technology; infrastructure and environmental constraints; fragmented and inaccessible data; and talent retention and commercialization. The number of assigned challenges ranged from zero to nine. We provided these challenges to the agent alongside the persona description and the policy prompt as context that structured, rather than determined, the agent's response.

\begin{figure}[h]
\centering
\small
\begin{tabular}{lp{0.72\linewidth}}
\toprule
Agent & Assigned landscape challenges \\
\midrule
PIAAC-4362 & Low public trust and AI literacy \\
PIAAC-1754 & Low public trust and AI literacy; Unequal access and disproportionate harms; Infrastructure and environmental constraints; Talent retention and commercialization; Privacy, discrimination, and online harms \\
PIAAC-10565 & Fragmented and inaccessible data; Privacy, discrimination, and online harms; Dependence on foreign technology; Unequal access and disproportionate harms; Infrastructure and environmental constraints; Talent retention and commercialization; Low public trust and AI literacy; Employment disruption \\
\bottomrule
\end{tabular}
\caption{Examples of the landscape challenges assigned to different agents.}
\label{fig:challenges}
\end{figure}

\subsection{Prompt Design for Single-Agent Evaluation}\label{app:prompts}
For each rating, the agent provided a brief justification and selected a value from 1 to 6. Lower values indicated opposition, low effectiveness, or limited benefit, and higher values indicated support, effectiveness, or substantial benefit. Because the rating questions appeared in a randomized order, we stored the realized order so that each response could be mapped back to the correct rating dimension.

The agents were given the web search skill, which allowed them to support externally verifiable factual claims with sources. The prompt also instructed each agent to report its information effort as none, one source, or several sources. This field was intended to reflect whether a person with the persona's background, skills, and temperament would realistically seek information about a government policy before forming an opinion. The agents were also instructed to avoid personal details that are not in their PIAAC-derived descriptions. For example, a persona with lower reported digital technology use might reasonably express concern about access to AI training. However, the agent should not turn that concern into an invented fact, such as claiming that the person had never used a computer or had been excluded from a particular service.

\subsection{Dyad Selection}\label{app:dyads}
For each policy, the number of possible dyads was the product of the numbers of lower- and higher-rating agents. The 11 policies without possible dyads lacked either agents with consistently low ratings or agents with consistently high ratings. The distance $D$ identified the maximum rating distance and the total number of possible contrasting dyads for each policy. After filtering out mixed-language pairs, we ranked the dyads by $D$, i.e., the extent of their quantified disagreement. In the first step of the coverage-first procedure, we selected up to 10 high-distance dyads per eligible policy in a round-robin sequence, to ensure that policies with relatively few possible pairs were represented. In the second step, we filled the remaining sample with the dyads that had the highest $D$ across all policies. The 1,533 conversations covered nine of the 20 policies, each with five turns per speaker (15,330 agent turns in total). We verified for every conversation that one member had rated all three dimensions 1--3 and the other had rated all three 4--6 in the single-agent responses. The corpus is uneven: support for Canadian creators (933 conversations), small-business AI adoption (384), and postsecondary AI tools (131) account for 1,448 conversations (94.5\%).

\subsection{Rating Prediction Model}\label{app:model}
We paired each single-agent response with its support, perceived-effectiveness, and perceived-benefit ratings to form a labeled training dataset. We evaluated the fine-tuned XLM-RoBERTa model~\cite{conneau2020unsupervised} on held-out response--rating pairs using exact-match accuracy, macro-F1, mean absolute error, and quadratic weighted kappa~\cite{cohen1968weighted}. Because the ratings are ordinal, mean absolute error and weighted kappa distinguished small, adjacent-rating errors from larger disagreements. We also compared predicted and observed rating distributions using the Wasserstein distance~\cite{vallender1974calculation}. On the held-out pairs, the model achieved exact-match accuracy of 0.92, 0.86, and 0.87 for support, effectiveness, and benefit, respectively; quadratic weighted kappa of 0.84, 0.90, and 0.93; and mean absolute error of 0.08, 0.14, and 0.13. For comparison, a TF-IDF logistic-regression classifier trained on the same split reached accuracy of 0.84, 0.80, and 0.83.

\subsection{Dyad Corpus and Inferred Opinion Change}\label{app:corpus}
Since some policies had more representation in our dyad sample, counts of conversational themes are not equally sampled across policy domains, and a few low-rating personas recur in many pairings.

Because no conversation asked agents to report a new numerical rating, we inferred change from each speaker's first and last turns (3,066 speaker trajectories; Table~\ref{tab:opinion}). The TF-IDF logistic-regression classifier classified support as unchanged for 98.4\% of trajectories, whereas the fine-tuned XLM-RoBERTa model did so for 86.2\%. An LLM annotation of a 415-conversation subset (830 trajectories), which coded a single overall stance rather than the three ratings, instead marked 576 trajectories as higher and only 235 as unchanged. Both full-corpus models inferred more upward than downward movement on literacy effectiveness and direct benefit, but the sizes of these changes varied considerably, and we evaluated the models on held-out single-agent responses rather than on manually rated dialogue turns. Both models also assigned every first conversational turn an upper-half support rating, including turns from agents selected for lower-half support, and therefore did not recover the designed contrast.

\begin{table*}[!ht]
\caption{Inferred opinion change between first and last turns (percentages of speaker trajectories).}
\label{tab:opinion}
\small
\begin{tabular}{lrrrrrrr}
\toprule
 & \multicolumn{3}{c}{Support} & \multicolumn{2}{c}{Effectiveness} & \multicolumn{2}{c}{Benefit} \\
\cmidrule(lr){2-4}\cmidrule(lr){5-6}\cmidrule(lr){7-8}
Method & Same & Higher & Lower & Higher & Lower & Higher & Lower \\
\midrule
TF-IDF logistic regression & 98.4 & 0.9 & 0.6 & 9.2 & 1.3 & 48.7 & 12.0 \\
XLM-RoBERTa & 86.2 & 12.9 & 0.8 & 30.6 & 4.8 & 36.2 & 8.3 \\
\bottomrule
\end{tabular}
\end{table*}

\subsection{LLooM Analysis}\label{app:lloom}
Following social computing research guidelines~\cite{mcdonald2019reliability}, we did not calculate inter-rater reliability for the automatically generated concepts and instead used them as prompts for interpretation. Concepts that emerged for the single-agent responses included access, training, trust, privacy, safety, equity, and institutional capacity. The qualitative analysis aimed to identify recurring arguments, concerns, implementation conditions, and interaction patterns that the rating analysis could not capture. For the single-agent responses, the analyzed texts captured how agents interpreted each policy, why they supported or opposed it, how effective they expected it to be, and whether they believed it would benefit them directly. For the dyads, we identified sources of disagreement, persistent concerns, shared positions, and qualifications, and the analysis also examined how subsequent turns accepted, challenged, qualified, or extended earlier points, and identified counterarguments and other concepts related to apparent movement toward or away from the other agent's position. In doing so, it treated implementation measures and proposed solutions as products of interaction, rather than as isolated statements.

\subsection{Visualization and Reproducibility}\label{app:repro}
\begin{figure*}[!ht]
    \centering
    \includegraphics[width=\linewidth]{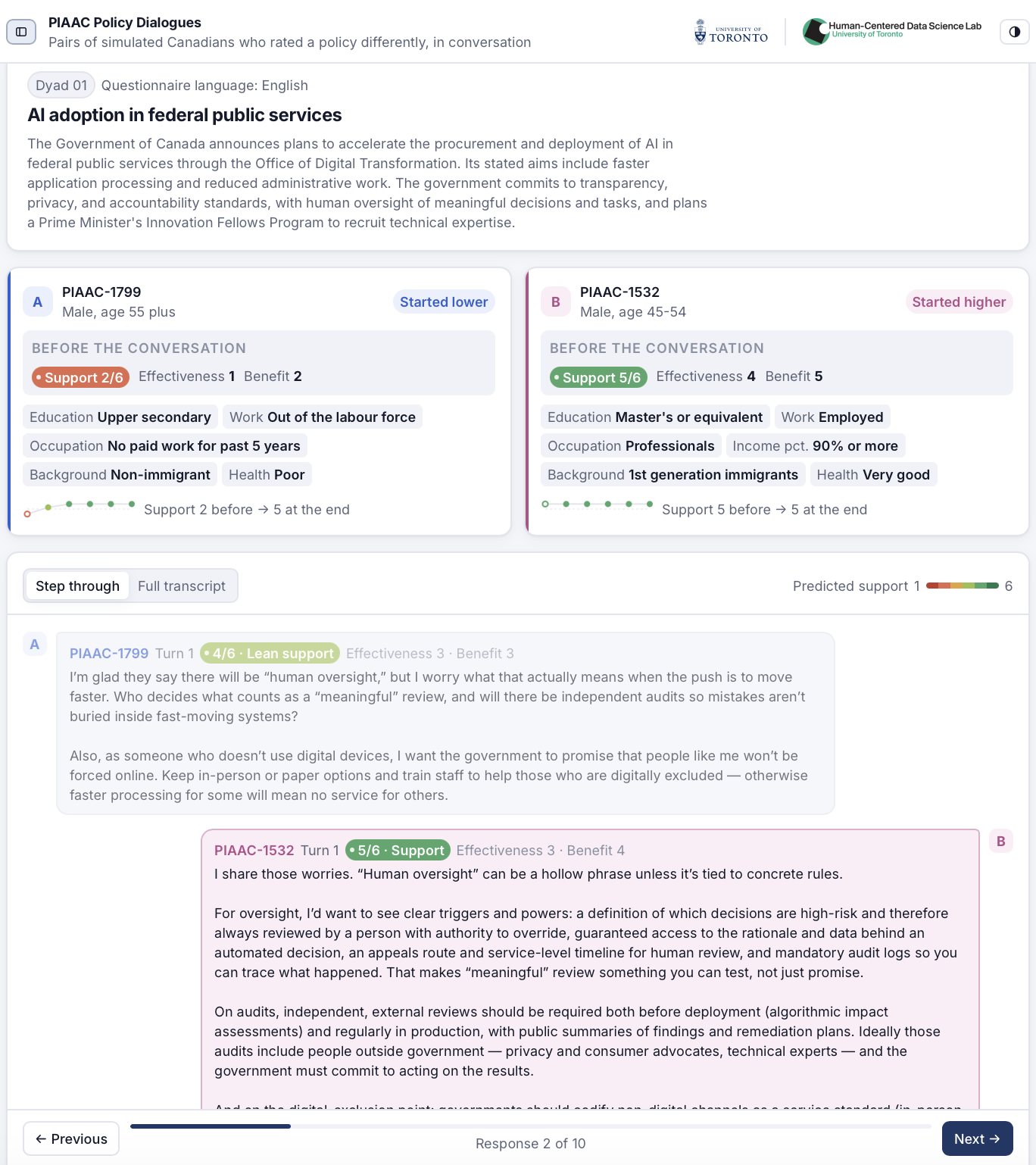}
    \caption{Dyad opinion viewer: one response at a time, but both agents remain visible.}
    \label{fig:dyad_viz}
\end{figure*}

We combined the conversation-level annotations with the original transcripts to produce a publicly accessible, interactive HTML viewer\footnote{\url{https://diptodas.github.io/piaac-dyad-viewer/}} (see Figure~\ref{fig:dyad_viz}). The viewer allows users to select a policy and a dyad, move through the conversation turn by turn, and inspect the position associated with each response, with supportive and opposing positions displayed in distinct colors. All randomized steps, including weighted sampling, challenge assignment, and response sampling, used fixed random seeds. We retained intermediate files (persona descriptions, policy responses, parsed workbooks, concept tables, sampled responses, annotations, and analysis manifests) to support traceability from the final findings back to the underlying model outputs.

\clearpage

\section{Findings Details}

\subsection{Additional Figures}\label{app:figures}
Here are the figures that offer additional (but not mandatory) details about the findings.

\begin{figure}[!ht]
\centering
\includegraphics[width=\linewidth]{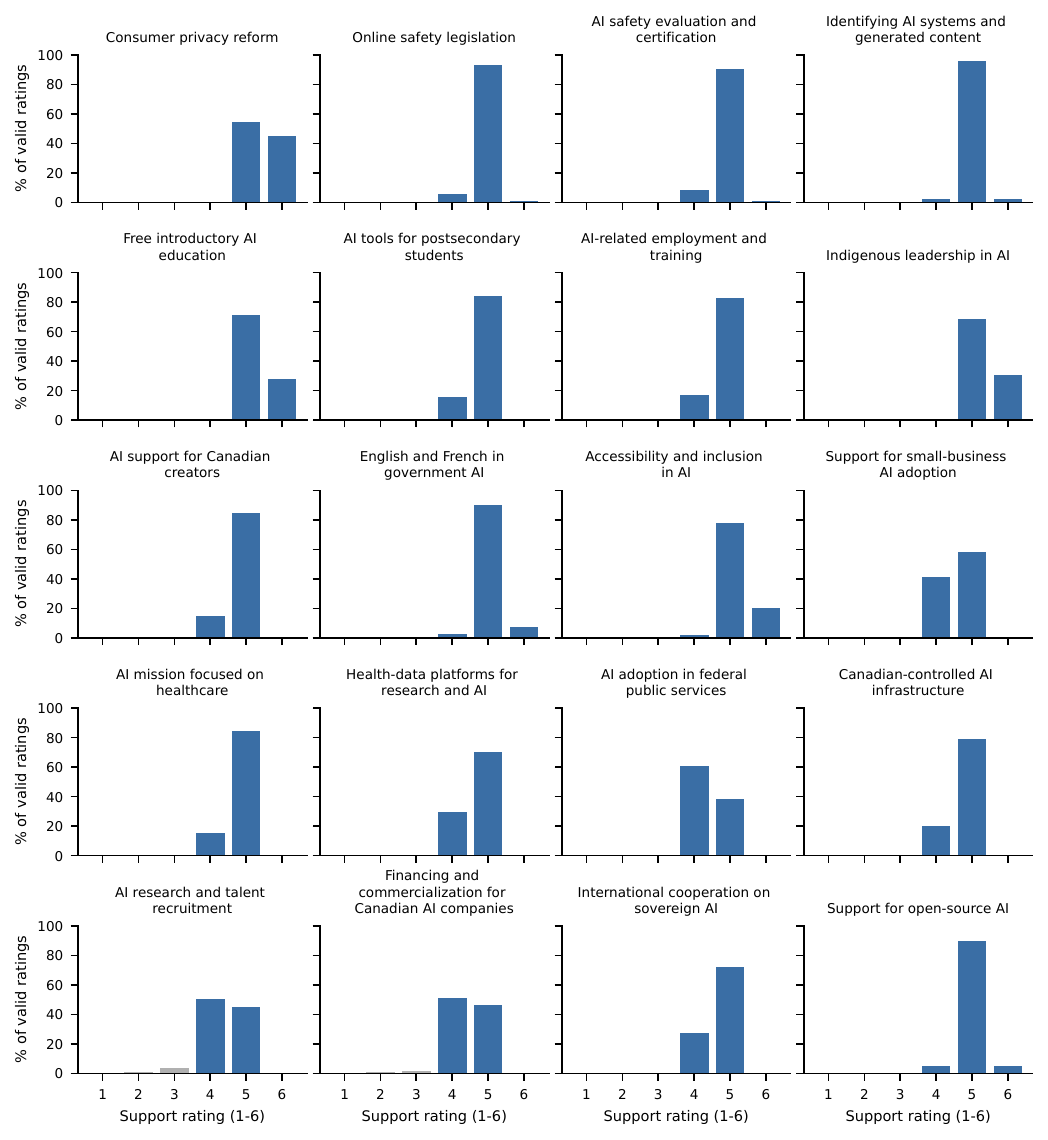}
\caption{PIAAC-persona-based agents' support for different AI policies. Bars show the percentage of valid ratings at each scale point; blue bars mark the upper half (4--6).}
\label{fig:support}
\end{figure}

\begin{figure}[!ht]
\centering
\includegraphics[width=\linewidth]{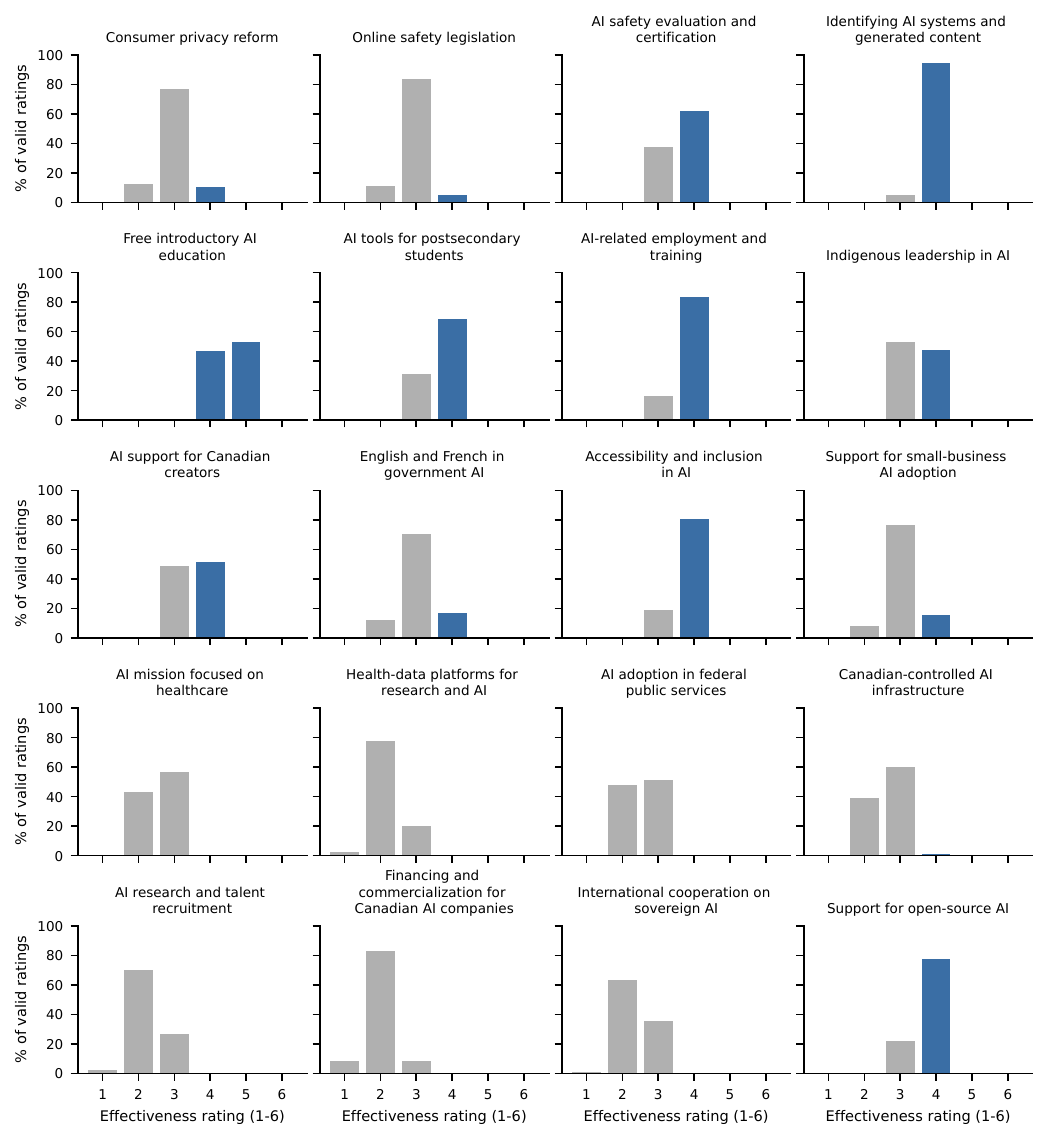}
\caption{PIAAC-persona-based agents' perceived effectiveness of different AI policies for improving adult AI literacy. Blue bars mark the upper half (4--6).}
\label{fig:effectiveness}
\end{figure}

\begin{figure}[!ht]
\centering
\includegraphics[width=\linewidth]{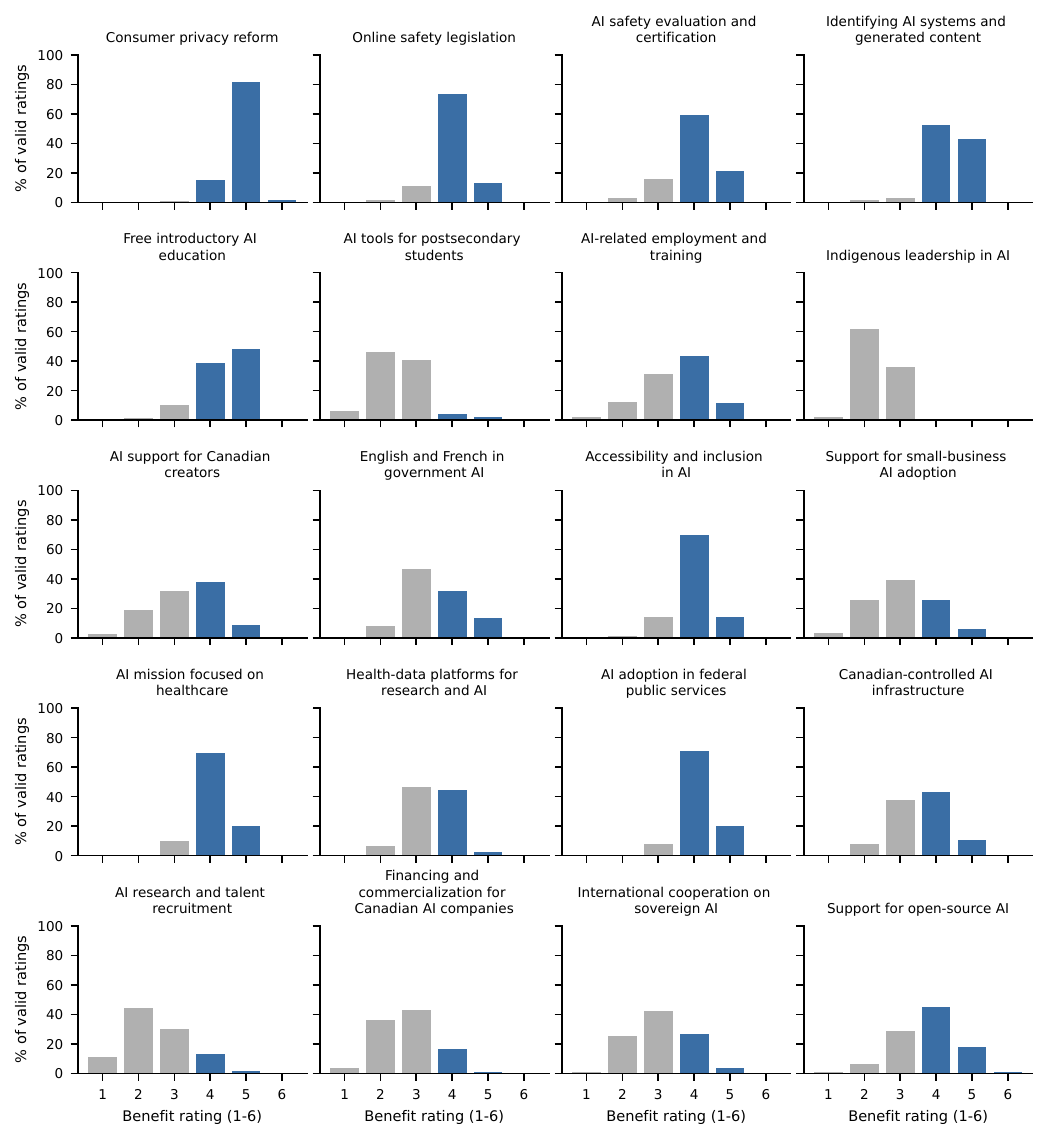}
\caption{PIAAC-persona-based agents' perceived direct benefit of different AI policies. Blue bars mark the upper half (4--6).}
\label{fig:benefit}
\end{figure}

\clearpage
\subsection{Additional Tables}\label{app:tables}
Here are the tables that offer additional (but not mandatory) details about the findings.

\begin{table*}[!ht]
\caption{Share of valid ratings in the upper half of the scale (4--6), by policy and dimension.}
\label{tab:upperhalf}
\small
\begin{tabular}{p{0.52\linewidth}rrr}
\toprule
Policy & Support (\%) & Benefit (\%) & Effectiveness (\%) \\
\midrule
All 20 policies & 99.5 & 56.4 & 35.9 \\
\midrule
Consumer privacy reform & 100.0 & 98.5 & 10.7 \\
Online safety legislation & 100.0 & 87.1 & 5.3 \\
AI safety evaluation and certification & 100.0 & 80.4 & 62.1 \\
Identifying AI systems and generated content & 100.0 & 95.2 & 94.9 \\
Free introductory AI education & 100.0 & 88.3 & 100.0 \\
AI tools for postsecondary students & 99.9 & 6.7 & 68.5 \\
AI-related employment and training & 100.0 & 54.9 & 83.4 \\
Indigenous leadership in AI & 100.0 & 0.4 & 47.3 \\
AI support for Canadian creators & 99.8 & 46.6 & 51.3 \\
English and French in government AI & 100.0 & 45.5 & 17.1 \\
Accessibility and inclusion in AI & 100.0 & 84.4 & 81.1 \\
Support for small-business AI adoption & 99.2 & 31.3 & 15.3 \\
AI mission focused on healthcare & 100.0 & 89.7 & 0.6 \\
Health-data platforms for research and AI & 99.8 & 46.9 & 0.3 \\
AI adoption in federal public services & 99.2 & 91.6 & 0.4 \\
Canadian-controlled AI infrastructure & 99.9 & 53.5 & 1.0 \\
AI research and talent recruitment & 95.3 & 14.5 & 0.2 \\
Financing and commercialization for Canadian AI companies & 97.5 & 17.3 & 0.1 \\
International cooperation on sovereign AI & 99.9 & 30.6 & 0.4 \\
Support for open-source AI & 100.0 & 64.2 & 78.0 \\
\bottomrule
\end{tabular}
\end{table*}

%TC:endignore
\end{document}